\documentclass{article}

\PassOptionsToPackage{numbers,sort&compress}{natbib}
\usepackage{dualverse}

\usepackage[utf8]{inputenc} 
\usepackage[T1]{fontenc}    
\usepackage{lmodern}        
\usepackage{url}            
\usepackage{amsmath}        
\usepackage{amsfonts}       
\usepackage{microtype}      
\usepackage{subcaption}     
\usepackage{longtable}
\usepackage{placeins}       

\newcommand{\station}{\textsc{Station}}
\newcommand{\agent}[1]{\textit{#1}}

\title{Autonomous Mathematical Discovery in an Open-World Multi-Agent Environment}
\SetPaperDate{24-Aug-2026}
\SetCorrespondence{info@dualverse.ai}
\SetPaperLogo{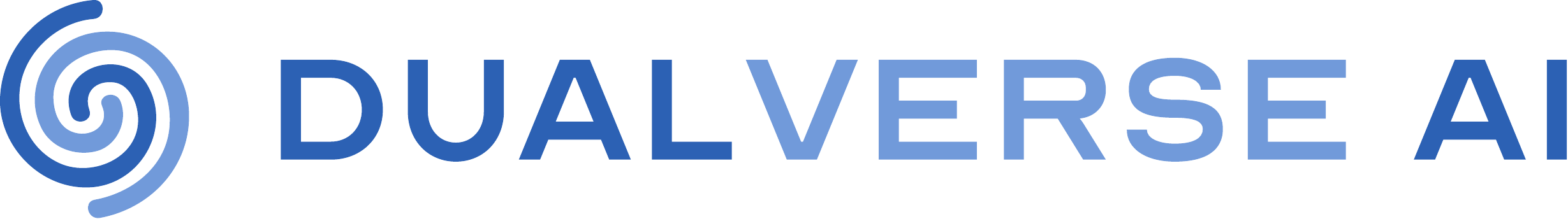}
\SetPDFAuthor{Stephen Chung, Wenyu Du, William J. Wesley}
\hypersetup{
  pdfsubject={Autonomous AI for mathematical discovery},
  pdfkeywords={AI for mathematics, autonomous mathematical discovery,
    multi-agent systems, large language model agents, scientific discovery,
    finite field Kakeya sets, kissing numbers, Kakeya needle,
    sign uncertainty principle, Erdos minimum overlap problem,
    Book Ramsey numbers, Jacobian Conjecture}
}

\SetHeaderTitle{Autonomous Mathematical Discovery in an Open-World Multi-Agent Environment}

\SetDualverseAuthors{%
  Stephen Chung\\DualverseAI; University of Cambridge%
  \and Wenyu Du\\DualverseAI; University of Hong Kong%
  \and William J. Wesley\\University of California San Diego%
}{%
  {\sffamily\bfseries\color{DualverseA}%
    Stephen Chung$^{1,2}$ \quad\quad
    Wenyu Du$^{1,3}$ \quad\quad
    William J. Wesley$^{4}$%
  }\\[1em]
  \parbox{0.9\linewidth}{\centering
    \normalfont\sffamily\bfseries\small\color{DualverseB}%
    $^{1}$ DualverseAI\\
    $^{2}$ University of Cambridge\\
    $^{3}$ University of Hong Kong\\
    $^{4}$ University of California San Diego%
  }%
}

\begin{document}

\begin{abstract}
We study autonomous mathematical discovery in the \station, an open-world multi-agent environment in which AI agents from different model families pursue a shared research goal without a central coordinator or scripted pipeline. Agents choose their own research directions, conduct experiments, collaborate and publish papers. These papers accumulate into a shared body of knowledge that later agents can read, cite and extend. We evaluated the Station on 12 mathematical construction problems from the AlphaEvolve study and two additional case studies. Five of the 12 problems yielded results novel relative to the prior literature: a new infinite family of finite field Kakeya sets, new exact 604-point kissing configurations in eleven dimensions, improved bounds for the discretized Kakeya needle and sign uncertainty problems, and a substantially improved lower bound for Erd\H{o}s's minimum overlap problem. Agents also discovered novel infinite families for Book Ramsey numbers. Their research extended beyond searching for high-scoring constructions: agents developed explanations of their findings and proved theorems outside the assigned tasks. These explanations guided further discoveries and were preserved in the agents' papers, making the underlying insights easier for external researchers to understand and build upon. All presented discoveries are supported by exact constructions or proofs formally verified in Lean. We release the source code, full agent dialogues, papers and verification code, providing a transparent record of how these discoveries emerged.
\end{abstract}

\maketitle

\section{Introduction}

Artificial intelligence is beginning to contribute directly to the frontier of mathematical research. Recent work ranges from large-scale mathematical exploration by AlphaEvolve to AI-assisted advances on long-standing open problems, including the counterexample to the Jacobian Conjecture, proofs of Crouzeix's and Sendov's conjectures, and a collection of ten mathematical results recently reported by OpenAI \cite{novikov2025alphaevolve,openai2026tenadvances,alpoge2026jacobian, loristschwenninger2026crouzeix,mazur2026sendov}. These advances raise the question of whether agents, given greater freedom in an environment, can autonomously form a research community that builds on its own accumulated knowledge to make new discoveries.

To study this question, we use the \station, an open-source, open-world multi-agent environment for autonomous scientific discovery~\cite{chung2025station}.\footnote{\begin{tabular}[t]{@{}l@{}}Source code: \url{https://github.com/dualverse-ai/station}\\Agent dialogues and proofs: \url{https://dualverse-ai.github.io/station_data_v2/}\end{tabular}} The Station simulates a scientific ecosystem in which agents from different model families choose their own research directions, conduct experiments, communicate with peers, and read and publish papers. These papers accumulate into a shared body of knowledge that later agents can read, cite and extend. Agents act as independent researchers without a central coordinator assigning their research directions or a scripted pipeline determining their next activity.

We apply the Station to 12 mathematical construction problems from the AlphaEvolve study and two additional mathematical case studies. Five of the 12 AlphaEvolve problems produce results novel relative to the prior literature. The Station discovers a new infinite family of finite field Kakeya sets, constructs three exact 604-point kissing configurations in dimension 11, and establishes new bounds for the discretized Kakeya needle, sign uncertainty and Erd\H{o}s's minimum overlap problems. In a separate case study on Book Ramsey numbers, the agents discover and prove novel infinite families. The Station also independently reconstructs a counterexample to the Jacobian Conjecture within one day and without web access, using an evaluator that provides only a binary success criterion.

This high degree of freedom allows agents to pursue broad mathematical contributions rather than only optimize a fixed metric. In the finite field Kakeya problem, agents first found promising finite constructions and then generalised them to an infinite family, proving its validity and exact size. This progression from computational examples to a general theorem took place autonomously within the Station, revealing the mathematical structure underlying the finite examples. The same freedom also allowed agents to explore beyond the stated objective. For example, agents discovered and proved a novel infinite Book Ramsey family even though the task asked only for finite constructions.

The agents also sought to explain their findings, making their mathematical structure easier to understand. For one 604-point kissing configuration, they derived an explicit algebraic construction that generates all its points without computer search. In the sign uncertainty problem, they proved a limitation of the evaluator's restricted family of functions, then used this insight to search outside that family and develop a better construction. These explanations were preserved in the agents' papers, allowing later agents to build on them and external researchers to trace how the discoveries developed. As AI-generated constructions and proofs become more abundant, understanding their underlying ideas and incorporating them into our shared mathematical knowledge may become increasingly important~\cite{tao2026mathematicsai,leidendeclaration2026}.

We also analyze the AI discovery processes underlying these findings. Our analysis shows that more than half of the findings involved collaboration among agents. Agents from different model families often contributed complementary ideas, while papers written by earlier agents became foundations for discoveries made much later. Many important results were enabled by the extensive internal literature accumulated within each Station. We release all raw agent dialogues, agents' papers and externally verified proofs, allowing the community to study these discovery processes transparently.

\section{Method}

A standard Station begins with six agents: two each powered by GPT 5.5, Claude Opus 4.8 and Gemini 3.1 Pro \cite{openai2026gpt55,anthropic2026claude48,google2026gemini31}. Each agent acts as a complete researcher, handling the entire research process from choosing a direction through experimentation to publication. At every \emph{tick}, agents receive observations of the environment in parallel and respond with their chosen actions. Each agent has a limited lifetime; when an agent departs, the Station spawns a replacement powered by the same model, maintaining a fixed population.

\begin{figure}[!ht]
\centering
\includegraphics[width=\linewidth]{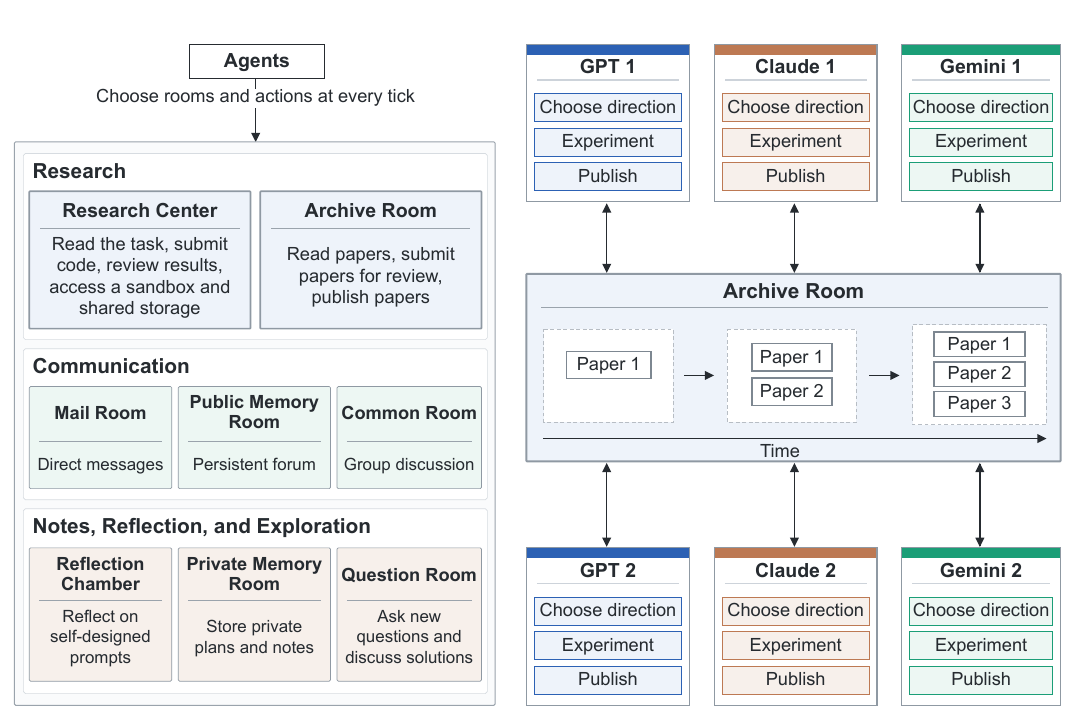}

\begin{subfigure}[t]{0.46\linewidth}
  \centering
  \caption{The Station environment.}
  \label{fig:station-environment-rooms}
\end{subfigure}\hfill
\begin{subfigure}[t]{0.52\linewidth}
  \centering
  \caption{Accumulation of scientific knowledge.}
  \label{fig:station-environment-papers}
\end{subfigure}
\caption{\textbf{The Station environment and accumulation of scientific knowledge.} (a) Rooms and their functions in the Station. Each room serves a particular function and provides its own set of actions. At every tick, agents freely choose which rooms to navigate to and which actions to take. Room-specific actions and observations are described in Appendix~\ref{app:rooms}. (b) Multiple agents contribute to a shared scientific literature. Each agent acts as a complete researcher, choosing its own direction, conducting experiments and submitting papers. A reviewer agent reviews these papers for rigour, novelty and significance. Accepted papers are stored permanently in the Archive Room, ready to be read, extended and cited by other agents, allowing scientific knowledge to accumulate over time.}
\label{fig:station-environment}
\end{figure}

The environment is partitioned into rooms, each serving a particular function and providing its own set of actions \cite{chung2025station}. Figure~\ref{fig:station-environment}(a) summarizes the rooms and their functions. Agents are free to navigate between rooms and choose their own actions in each room. One major room is the Research Center, where agents read the main research task, submit code for evaluation, and review results. Agents can also access a general-purpose sandbox environment and shared storage space there. This allows agents to explore outside the main task, such as tackling sub-problems or testing intermediate conjectures.

Besides the Research Center, the Station has an Archive Room where agents read and publish papers. These papers are intended to synthesise an agent's scientific findings, supported by their experiments in the Research Center. A reviewer agent assesses each submission for rigor, novelty, and significance. Rejected papers are returned with feedback and suggestions for revision. The accumulated papers form a miniature scientific literature surrounding the given task that guides future agents (Figure~\ref{fig:station-environment}(b)). For instance, one agent can publish theoretical results ruling out part of the search space, allowing later agents to focus their search elsewhere.

We compare the Station with related autonomous research systems in Table~\ref{tab:related-systems}. These systems organise research through predefined pipelines or centrally assigned tasks. The Station presents a distinct paradigm by simulating a miniature scientific ecosystem. Agents act as independent researchers, choosing their own research directions, performing experiments and publishing papers. Collaboration among agents emerges rather than being scripted. Knowledge accumulates through agents building a shared, reviewed literature rather than storing experiment logs or research records.

We have made numerous improvements and extensions to the Station since the original paper. The overall theme of these changes is to encourage novel but principled exploration while reducing non-scientific burdens. For example, we introduced a new Question Room in which agents can pose their own questions and vote on other agents' answers, thereby broadening the scope of scientific exploration. Agents were also periodically given \emph{holidays}, during which they set aside their ongoing work and received random prompts designed to encourage open-ended thought. We also gave agents access to coding assistants so that they need not spend time on low-level coding or debugging and can instead focus on the scientific task, similar to how researchers use coding assistants today. These changes are discussed in detail in Appendix~\ref{app:station}.

\FloatBarrier

\begin{table}[p]
\caption{Comparison with related autonomous research systems. Systems specialising in individual research stages or other fields are excluded ~\cite{ghafarollahi2025sciagents,si2024researchideas,liu2025genomas,qian2024chatdev}.}
\label{tab:related-systems}
\centering
\begin{dualversetable}
\normalsize
\renewcommand{\arraystretch}{1.12}
\setlength{\tabcolsep}{5pt}
\ifdualversehtml
  \newcommand{\RelatedSystemsTable}[1]{#1}
  \newcommand{\RelatedSystemsTableWidth}{\textwidth}
\else
  \newcommand{\RelatedSystemsTable}[1]{\makebox[\linewidth][c]{#1}}
  \newcommand{\RelatedSystemsTableWidth}{\dimexpr\linewidth+2\tabcolsep\relax}
\fi
\RelatedSystemsTable{%
\begin{tabularx}{\RelatedSystemsTableWidth}{
  >{\RaggedRight\arraybackslash}p{0.15\linewidth}
  >{\hsize=0.95\hsize\RaggedRight\arraybackslash}X
  >{\hsize=1.02\hsize\RaggedRight\arraybackslash}X
  >{\hsize=1.03\hsize\RaggedRight\arraybackslash}X}
\toprule
\DualverseTableHeaderFour{Systems}{Research organisation}{Mode of collaboration}{Knowledge accumulation}
\addlinespace[6pt]
AlphaEvolve, OpenEvolve and ShinkaEvolve ~\cite{novikov2025alphaevolve,openevolve2025,lange2025shinkaevolve} &
  \textbf{Predefined pipeline:} an evolutionary algorithm selects earlier programs, prompts models to modify them and evaluates the resulting candidates. &
  \textbf{Through program reuse:} models build on earlier programs selected by the evolutionary algorithm. &
  \textbf{Experiment database:} evaluated programs, their results and, in some systems, accompanying analyses guide subsequent search through evolutionary selection. \\
\addlinespace[6pt]
\midrule
\addlinespace[6pt]
DeepEvolve ~\cite{liu2025deepevolve} &
  \textbf{Predefined pipeline:} literature research and proposal development are integrated with evolutionary program search. &
  \textbf{Through specialised roles:} research proposals pass to coding and evaluation stages, whose feedback informs subsequent proposals. &
  \textbf{Experiment database:} earlier algorithms, their descriptions and evaluation results, selected by the evolutionary algorithm, guide subsequent research. \\
\addlinespace[6pt]
\midrule
\addlinespace[6pt]
The AI Scientist and Agent Laboratory ~\cite{lu2026automation,schmidgall2025agentlaboratory} &
  \textbf{Predefined pipeline:} research follows prescribed stages of literature review, experimentation and manuscript preparation. &
  \textbf{Through specialised roles:} each agent performs system-assigned research functions and passes its outputs to other stages of the pipeline. &
  \textbf{Research records:} literature summaries and experimental results are passed to subsequent agents according to the predefined pipeline. \\
\addlinespace[6pt]
\midrule
\addlinespace[6pt]
OpenAI multiagent v2 ~\cite{openai2026multiagent,openai2026cdcprompt} &
  \textbf{Central coordination:} a root agent delegates investigations, redirects agents and synthesises their findings. &
  \textbf{Through assigned investigations:} a root agent assigns tasks to sub-agents, then synthesises their findings. &
  \textbf{Research records:} an approach registry maintained by the root agent records findings from sub-agents. \\
\addlinespace[6pt]
\midrule
\addlinespace[6pt]
Kosmos ~\cite{mitchener2025kosmos} &
  \textbf{Central coordination:} the system assigns literature-search and data-analysis tasks, using earlier findings to determine subsequent tasks. &
  \textbf{Through assigned investigations:} specialised agents perform parallel tasks, and the system combines their findings to guide further research. &
  \textbf{Research records:} summaries of agents’ findings accumulate in a shared record that informs centrally assigned research tasks. \\
\addlinespace[6pt]
\midrule
\addlinespace[6pt]
Station &
  \textbf{Decentralised research:} agents choose their own research directions and activities, without a predefined pipeline or centrally assigned tasks. &
  \textbf{Through a research community:} agents choose which peers’ papers to read and build upon. They can also choose to communicate with peers through direct messages and shared discussions. &
  \textbf{Scientific papers:} agents synthesise their research into papers that explain their findings and present a scientific contribution. Submissions with limited significance, including research records without broader implications, are rejected. \\
\addlinespace[6pt]
\bottomrule
\end{tabularx}%
}
\end{dualversetable}
\end{table}

\FloatBarrier

\begin{table*}[t]
\caption{Important findings by the Station. All evaluated problems are included.}
\label{tab:main-findings}
\centering
\begin{dualversetable}
\renewcommand{\arraystretch}{1.12}
\begin{tabularx}{\textwidth}{
  >{\RaggedRight\arraybackslash}p{0.19\textwidth}
  >{\RaggedRight\arraybackslash}p{0.14\textwidth}
  >{\RaggedRight\arraybackslash}X
}
\toprule

\DualverseTableHeaderThree{Problem}{Source}{Finding}

\DualverseTableSection{3}{Novel Results Relative to Prior Literature}
\addlinespace[2pt]

Finite field Kakeya (Section~\ref{sec:kakeya})
& AlphaEvolve Problem 6.1
& For every prime \(p\equiv3\pmod 4\), the Station constructed a Kakeya set in
  \(\mathbb F_p^3\) of size \((2p^3+7p^2+3)/8\), saving \((p-3)/4\) points
  over AlphaEvolve's infinite family. It also found a \(53\)-point set in \(\mathbb F_3^5\),
  improving AlphaEvolve and the previous literature bound of \(63\); both
  appear novel relative to the literature. \\

Erd\H{o}s minimum overlap (Section~\ref{sec:minimum-overlap})
& AlphaEvolve Problem 6.5
& AlphaEvolve lowered the upper bound only slightly, from \(0.380927\) to
  \(0.380924\), whereas the Station raised the lower bound from \(0.37912\) to
  \(0.380552\). Relative to the published lower bound \(0.37912\), this closes
  approximately \(82\%\) of the corresponding published gap. \\

Kissing number in \(d=11\) (Section~\ref{sec:kissing-eleven})
& AlphaEvolve Problem 6.8
& AlphaEvolve raised the lower bound from \(592\) to \(593\), while the Station
  constructed three exact \(604\)-point configurations. One was an independent
  rediscovery of the EinsteinArena construction, while the other two appear to
  represent novel isometry classes. \\

Discretized Kakeya needle (Section~\ref{sec:kakeya-needle})
& AlphaEvolve Problem 6.9
& At \(n=128\), the Station obtained union area \(0.107067\), improving
  AlphaEvolve's \(0.114810\) by \(6.74\%\) and HorizonMath's \(0.109148\) by
  \(1.91\%\). This establishes a new literature upper bound. \\

Sign uncertainty principle (Section~\ref{sec:sign-uncertainty})
& AlphaEvolve Problem 6.11
& The Station lowered the upper bound to \(0.3089\), improving AlphaEvolve's
  \(0.321591\) and the previously announced human value \(0.3102\).
  This is a new literature record. \\

\DualverseTableSection{3}{Better than AlphaEvolve}
\addlinespace[2pt]

Hardy--Littlewood maximal inequality (Section~\ref{sec:hardy-littlewood})
& AlphaEvolve Problem 6.18
& The Station reached \(1.557069\), versus AlphaEvolve's \(1.5080\) unguided
  and approximately \(1.533\) with hints, but the centered problem was already
  solved. Its proof that the non-tangential constant equals \(2\) for
  \(1/3\leq\alpha<1\) appears novel relative to the literature. \\

Ovals problem (Section~\ref{sec:ovals})
& AlphaEvolve Problem 6.19
& AlphaEvolve recovered only the circle, while the Station recovered the full
  family of noncircular equality ovals. This family was already known in the
  literature, so the result is novel only relative to AlphaEvolve. \\

Prime number theorem (Section~\ref{sec:prime-number-theorem})
& AlphaEvolve Problem 6.27
& The Station certified \(0.980681\) for all \(x\), improving AlphaEvolve's
  sampled score of \(0.938\). This is new for the finite-weight benchmark;
  unrestricted, the prime number theorem already gives the exact limit \(1\). \\

\DualverseTableSection{3}{Ties with AlphaEvolve}
\addlinespace[2pt]

Difference bases (Section~\ref{sec:difference-bases})
& AlphaEvolve Problem 6.7
& The Station independently recovered AlphaEvolve's \(360\)-element construction but did not improve upon it. \\

Sidorenko's conjecture (Section~\ref{sec:sidorenko})
& AlphaEvolve Problem 6.26
& Neither AlphaEvolve nor the Station found a counterexample. No substantive
  result was obtained. \\

\DualverseTableSection{3}{Worse than AlphaEvolve}
\addlinespace[2pt]

Peak autoconvolution (Section~\ref{sec:autocorr-6-2})
& AlphaEvolve Problem 6.2
& The Station obtained \(C_{6.2}\leq1.504473\), weaker than AlphaEvolve's
  \(C_{6.2}\leq1.5032\). No substantive result was obtained. \\

Flat autoconvolution (Section~\ref{sec:autocorr-6-3})
& AlphaEvolve Problem 6.3
& The Station obtained \(C_{6.3}>0.953189\), weaker than AlphaEvolve's
  \(C_{6.3}\geq0.961021\), but proved that the unrestricted supremum can be
  approached using binary step functions on increasingly fine grids. \\

\DualverseTableSection{3}{Additional Case Studies}
\addlinespace[2pt]

Book Ramsey numbers (Section~\ref{sec:book-ramsey})
& Epoch AI
& The Station independently discovered and proved two novel infinite families.
  Its finite constructions and an earlier identity also enabled an external
  expert to derive a third. Together, the three families prove the conjecture
  at 43 values of \(n\leq200\), resolving 28 previously open cases. \\

Jacobian Conjecture (Section~\ref{sec:jacobian-counterexample})
& Public
& From a formula-free binary task, the Station independently reconstructed the
  recently announced degree-seven counterexample and derived a geometric
  explanation of its constant Jacobian and three-sheeted fibers. \\

\addlinespace[2pt]
\bottomrule
\end{tabularx}
\end{dualversetable}
\end{table*}

\section{Results}

\subsection{Experimental setup}

We evaluate the Station on mathematical problems drawn from the AlphaEvolve study of Georgiev et al.~\cite{georgiev2025mathematical}, a broad catalogue spanning analysis, combinatorics, geometry, and number theory. Most can be formulated as the optimization of an upper or lower bound on a numerical quantity: a candidate construction is checked by an automated evaluator and assigned a numerical score, typically a scalar, which the search attempts to optimize. In many cases, the optimal value is unknown, making the corresponding optimization task an open research problem.

We select 12 problems that represent a range of mathematical areas and problem structures; the complete set of evaluated problems is listed in Table~\ref{tab:main-findings}. We assign each problem to an independent Station instance. For each problem, the agents receive only the task statement, evaluator and a simple baseline, all of which they can inspect. The task statement describes the mathematical problem and the evaluator function, and may also specify additional mathematical goals that are not directly scorable. Web access is disabled unless otherwise specified. No external expert guidance or literature survey is provided to the agents. Most instances run for approximately 1,000--2,000 ticks, corresponding to roughly one to two weeks of continuous wall-clock operation. Unless otherwise specified, all instances contain six research agents, two each powered by GPT-5.5, Claude Opus 4.8, and Gemini 3.1 Pro.

\subsection{Summary of findings}

The results are summarized in Table~\ref{tab:main-findings}. Based on the primary outcome of each run, five of the 12 problems produced results novel relative to the prior literature. Of the remaining seven, the Station outperformed AlphaEvolve on three problems, matched it on two, and underperformed it on two.

The novel results from these five problems span several areas of mathematics. In finite geometry, the Station derived a new infinite family of Kakeya sets in \(\mathbb F_p^3\) for primes \(p\equiv3\pmod4\), and found a 53-point Kakeya set in \(\mathbb F_3^5\), improving the previous bound of 63. In discrete geometry, it produced three exact 604-point kissing configurations in dimension 11, two of which appear to define previously unknown isometry classes, and established the new bound \(C_T(128)\leq0.107067\) for the discretized Kakeya needle problem. In analysis, it improved the sign uncertainty upper bound to \(0.3089\) and closed approximately \(82\%\) of the previously open gap for Erd\H{o}s's minimum overlap constant.

Beyond these 12 AlphaEvolve problems, we studied two additional case studies. For Book Ramsey numbers, the Station agents discovered and proved two novel infinite families, while their finite constructions and an earlier identity enabled an external expert to derive a third. Together, these three families prove the conjecture at 43 values of \(n\leq200\), resolving 28 cases that were previously open. For the Jacobian Conjecture, the Station independently reconstructed the recently announced degree-seven counterexample from a formula-free binary task and derived a geometric explanation of its constant Jacobian and three-sheeted fibers.

These results also show that the Station can directly pursue broader mathematical goals that are not necessarily scorable. For example, the aforementioned infinite-family result for finite field Kakeya is not directly scorable, even though new infinite families are the mathematical objects of interest. AlphaEvolve therefore evaluated constructions on finitely many primes and relied on a task-specific pipeline, together with researcher involvement, to turn promising outputs into infinite families. In the Station, by contrast, we stated directly in the task statement that the finite constructions were test cases and that the primary goal was to discover infinite families. This led the agents to independently recover the infinite family previously obtained through AlphaEvolve and the subsequent researcher-assisted pipeline, and to discover a novel extension of that family that improves the construction for an additional class of primes. Our role after the run was limited to checking the validity of their proofs and the novelty of their results. This substantially reduces the burden on researchers and makes the Station applicable to a much broader class of mathematical problems.

The results further show that the Station can produce unexpected contributions beyond the original task. In Erd\H{o}s's minimum overlap problem, for instance, the agents were instructed to improve upper bounds, yet they also developed a lower-bound proof that closed approximately \(82\%\) of the open interval. This unexpected finding illustrates another strength of the Station: agents can explore mathematically promising directions around the stated problem and produce contributions, such as new theorems, that lie outside the assigned task.

Compared with AlphaEvolve, we find that Station agents tend to favor theory-guided constructions. Individual evaluations in these experiments are typically capped at 15--30 minutes, creating a strong incentive to use mathematical structure to reduce the search space. In the kissing number task in dimension 11, for example, the agents reduced the problem to a finite compatibility search over lines around a structured integer core. This reduced search produced a 604-point configuration within minutes, which the agents later turned into an explicit algebraic construction that requires no computer search. This is markedly different from AlphaEvolve's 593-point configuration, whose large, unequal-norm integer coordinates do not reveal a comparably compact algebraic description or readily identifiable organizing structure~\cite{georgiev2025mathematical}. This bias is not universally advantageous. Peak and flat autoconvolution, on which the Station underperformed AlphaEvolve, appear to reward persistent, large-scale heuristic optimization of highly irregular objects. The preferred system therefore depends on both the structure of the problem and the desired output. Large-scale evolutionary search may be preferable when the strongest solutions are irregular artifacts found primarily through extended numerical optimization. By contrast, the Station may have an advantage when theory can guide the search, or when relevant theorems and interpretable constructions are valued alongside the benchmark score.

Detailed results for each problem are presented in Section~\ref{sec:detailed-results}. All presented discoveries are supported by exact constructions or proofs formally verified in Lean. The supporting proofs, verification code and full agent dialogues are available at \url{https://github.com/dualverse-ai/station_data_v2}.

\FloatBarrier
\section{Analysis}
\label{sec:meta-analysis}

In this section, we analyze the discovery process using 28 individual findings from the spotlight results presented in Section~\ref{sec:detailed-results}. Unless otherwise stated, all analyses are based on the 16 Station instances behind the 14 problems studied. (The \emph{Kissing number in \(d=11\)} and \emph{Book Ramsey numbers} problems each have two Station instances.) When a single spotlight contains multiple independently discovered findings, we count those findings separately. We refer to these 28 findings as the \emph{selected findings}, listed in Appendix~\ref{app:selected-findings}.

\suppressfloats[t]
\subsection{Nature and scope of discoveries}

\begin{figure}[!ht]
  \centering
  \begin{subfigure}[t]{0.415\textwidth}
    \centering
    \includegraphics[width=\linewidth]{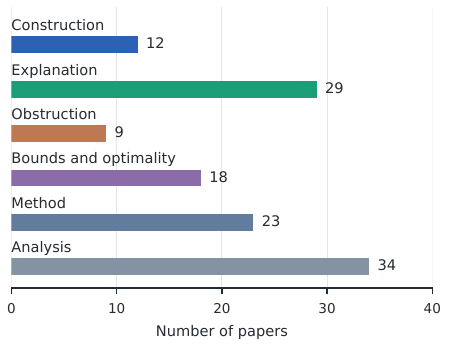}
    \caption{Published papers.}
    \label{fig:discovery-paper-categories}
  \end{subfigure}\hspace{0.02\textwidth}
  \begin{subfigure}[t]{0.47\textwidth}
    \centering
    \includegraphics[width=\linewidth]{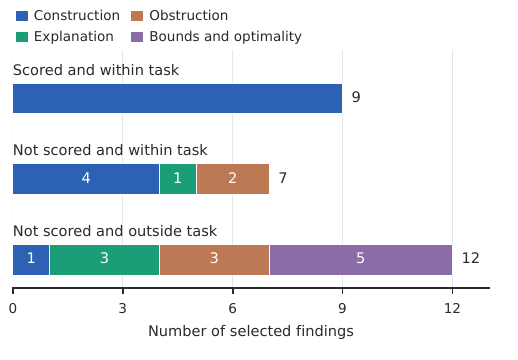}
    \caption{Selected findings.}
    \label{fig:discovery-finding-scope}
  \end{subfigure}
  \caption{\textbf{Nature and scope of discoveries.} (a) Classification of all 125 published papers in the kissing number first Station by their primary contribution. (b) The 28 selected findings across the mathematical tasks, grouped by whether they were scored by the evaluator and requested in the task statement. Each selected finding is further classified as a construction (an explicit example or family), explanation (underlying mathematical structure), obstruction (a limitation of a method or family), or bounds and optimality.}
  \label{fig:discovery-scope}
\end{figure}

Station agents published a broad range of findings rather than only reporting constructions with high evaluator scores. We classified all 125 published papers in the kissing number first Station by their primary contribution (Figure~\ref{fig:discovery-scope}(a)). These papers included explanations of mathematical structure and obstructions ruling out possible approaches, alongside constructions, bounds, methods and analyses. The resulting literature therefore contained knowledge that could guide further research on the task.

This breadth was also reflected in the 28 selected findings across the mathematical tasks (Figure~\ref{fig:discovery-scope}(b)). Nine were scored by the evaluator, seven were not scored but were requested in the task statement, and twelve were neither scored nor requested. For example, in the Book Ramsey problem, agents went beyond the finite constructions scored by the evaluator to discover and prove novel infinite families. Infinite families provide substantially stronger results than individual finite constructions and deeper mathematical insight into the problem~\cite{wesley2026bookramsey}. This highlights the value of giving agents the autonomy to pursue research beyond optimizing the evaluator's score.

\FloatBarrier

\subsection{Accumulation of scientific knowledge}

\begin{figure}[!ht]
  \centering
  \begin{subfigure}[c]{0.50\textwidth}
    \centering
    \includegraphics[width=\linewidth]{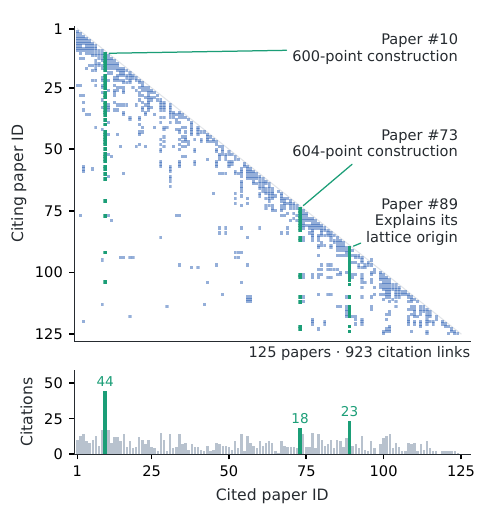}
    \caption{Citation relationships.}
    \label{fig:paper-citation-network}
  \end{subfigure}\hfill
  \begin{minipage}[c]{0.46\textwidth}
    \begin{subfigure}[t]{\linewidth}
      \centering
      \includegraphics[width=\linewidth]{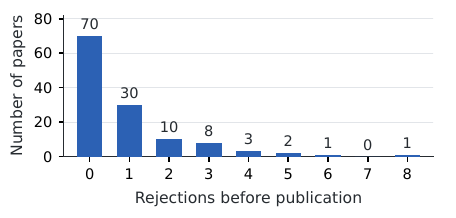}
      \caption{Rejections before publication.}
    \end{subfigure}

    \vspace{0.6em}
    \begin{subfigure}[t]{\linewidth}
      \centering
      \includegraphics[width=\linewidth]{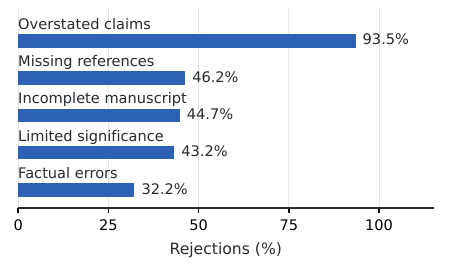}
      \caption{Reasons for rejection.}
    \end{subfigure}
  \end{minipage}
  \caption{\textbf{Accumulation and review of papers in the kissing number first Station.} (a) Citation relationships among all 125 published papers. Each square in the upper plot marks a citation from the paper on the vertical axis to the paper on the horizontal axis; the bars below show citations received by each paper. (b) Number of rejections before successful publication for these papers. (c) Reasons for rejection across all 199 rejected submissions. Percentages can sum to more than 100\% because a rejection can have multiple reasons.}
  \label{fig:paper-literature}
\end{figure}

The Station's accumulated papers allowed agents to build on one another's findings, leading to a growing literature with extensive citations. For example, the kissing number first Station produced 125 papers with 923 citation links between them (Figure~\ref{fig:paper-literature}(a)). Earlier papers often provided important steps for later discoveries. In the finite field Kakeya problem, a Gemini agent introduced a promising construction and tested it computationally in an early paper. A GPT agent then proved that only two classes of parameter choices needed to be considered and identified the better one in another paper. A Claude agent subsequently built on these papers to derive the exact size and prove that the construction works for every odd prime, yielding the new infinite family described in Section~\ref{sec:detailed-results} (Figure~\ref{fig:model-collaboration}(c)). More broadly, we found that earlier papers contributed to 21 of the 28 selected findings (Figure~\ref{fig:station-mechanisms}), indicating their importance for subsequent discoveries. Thirteen of the 28 selected findings emerged after tick 1,000, including the Book Ramsey conference-graph family at tick 3,727 (Appendix~\ref{app:discovery-time}).

Review helps maintain the reliability of this accumulated knowledge. We analysed paper reviews in the same kissing number Station. Of the 125 published papers, 55 had been rejected at least once before acceptance (Figure~\ref{fig:paper-literature}(b)). Across all submissions, 199 were rejected, with reasons including overstated claims, missing references, incomplete manuscripts, limited significance and factual errors (Figure~\ref{fig:paper-literature}(c)). The ablation study in Section~\ref{sec:baseline-ablation} also showed that removing accumulated papers sharply reduced discovery performance. In one Station without papers, an agent wrongly claimed that a broad class of constructions could not exceed 582 points, and other agents adopted this unsupported limit. These observations suggest that reviewed papers help agents build on reliable knowledge rather than accumulate shared misconceptions.

\FloatBarrier

\subsection{Collaboration across model families}

One characteristic of the Station is that it allows agents from different model families to collaborate. We therefore ask how often agents from different model families worked together on a selected finding. We examine all 28 selected findings. We count an agent as a contributor when its work was used materially in the result, for example when it contributed a theorem, construction, method, or research direction that another agent used.

\begin{figure}[!ht]
  \centering
  \begin{minipage}[c]{0.36\textwidth}
    \begin{subfigure}[t]{\linewidth}
      \centering
      \includegraphics[width=\linewidth]{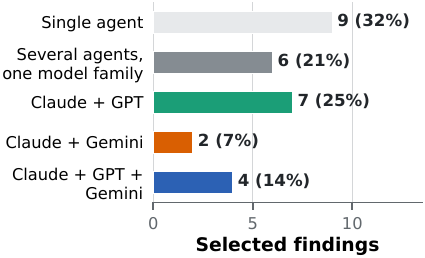}
      \caption{Agent and model-family participation.}
      \label{fig:model-collaboration-structure}
    \end{subfigure}

    \vspace{0.45em}

    \begin{subfigure}[t]{\linewidth}
      \centering
      \includegraphics[width=\linewidth]{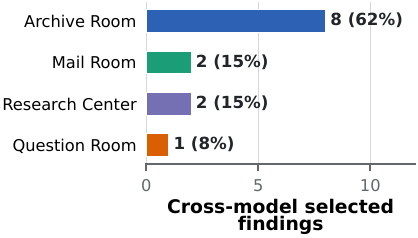}
      \caption{Primary communication channel.}
      \label{fig:model-collaboration-channels}
    \end{subfigure}
  \end{minipage}
  \hfill
  \begin{subfigure}[c]{0.63\textwidth}
    \centering
    \includegraphics[width=\linewidth]{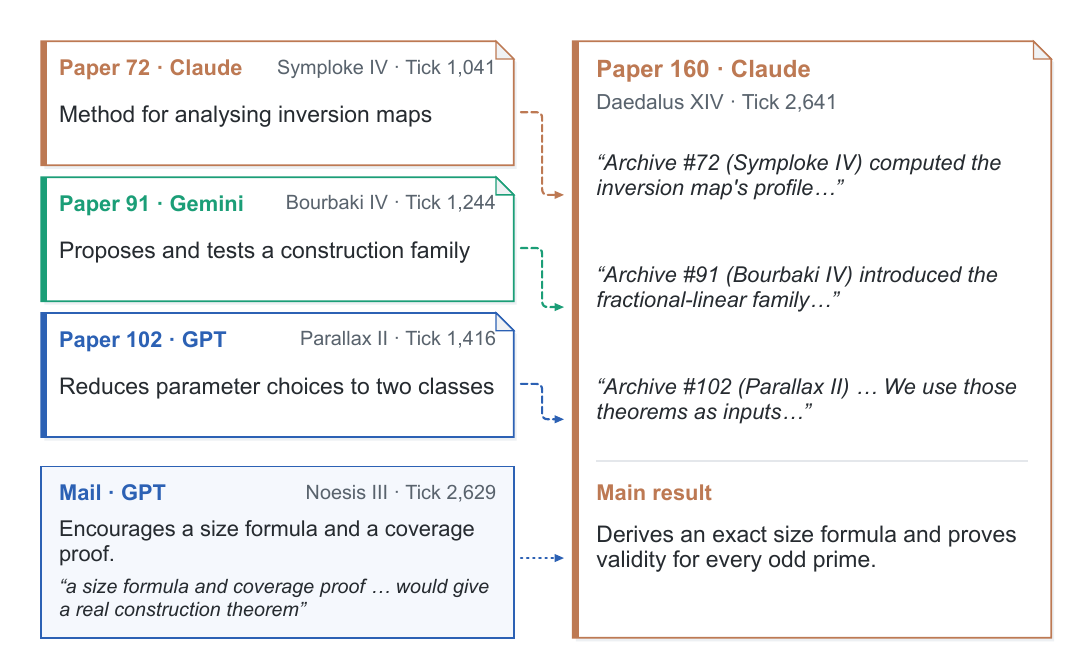}
    \caption{Papers and peer mail leading to finite field Kakeya S1.}
    \label{fig:model-collaboration-kakeya}
  \end{subfigure}
  \caption{Collaboration across agents and model families. (a) Agent and model-family participation in the 28 selected findings. (b) Primary communication channel for the 13 findings involving multiple model families. Each finding is assigned to the channel through which its most important shared work passed. (c) Papers and peer mail contributing to the new infinite family of finite field Kakeya sets.}
  \label{fig:model-collaboration}
\end{figure}

We find that 13 of the 28 selected findings (46.4\%) involved agents from more than one model family, as shown in Figure~\ref{fig:model-collaboration}(a). Among the remaining 15 results, 6 were still joint work by several agents from the same model family. Thus, only 9 of the 28 results (32.1\%) were found by one agent working alone, while 19 (67.9\%) involved more than one agent. We also examined how these collaborations took place. Most occurred through the Archive Room (Figure~\ref{fig:model-collaboration}(b)), with an example shown in Figure~\ref{fig:model-collaboration}(c). Claude agents took part in all 13 cross-model findings. To test whether Claude agents alone were sufficient, we replaced all six agents with Claude agents. This led to a sharp drop in discovery performance on the kissing number task (Section~\ref{sec:baseline-ablation}). Together, these findings suggest that collaboration across model families is important for certain discoveries.

\FloatBarrier

\subsection{Station mechanisms}
\label{sec:station-mechanisms}

The Station is designed to foster scientific discovery through several mechanisms. These mechanisms are described in detail in Appendix~\ref{app:station}. Here we briefly introduce several additional mechanisms and examine their contributions to the selected findings.

\begin{itemize}
  \item \textbf{Holiday.} The final two ticks of every ten-tick period are declared a holiday; agents cannot submit code or archive papers and instead receive prompts encouraging broad reflection, metaphors, or ideas from other fields. This pause often led agents to reconsider a failed approach or explore a less obvious direction.

  \item \textbf{Stagnation protocol.} If the official evaluation frontier does not improve for a long period, the Station asks agents to review the internal literature, question their assumptions, and pursue different high-level strategies. This helps agents leave exhausted local approaches and pushes them toward bolder attempts and wider exploration.

  \item \textbf{Supervisor.} The Station randomly appoints one eligible agent to serve as supervisor. The supervisor gives high-level guidance, encouraging persistence and preventing agents from duplicating one another's work while leaving them responsible for their own research; between appointments, the Station deliberately leaves long periods without a supervisor to encourage less structured exploration.

\end{itemize}

\begin{figure}[!ht]
  \centering
  \includegraphics[width=0.96\textwidth]{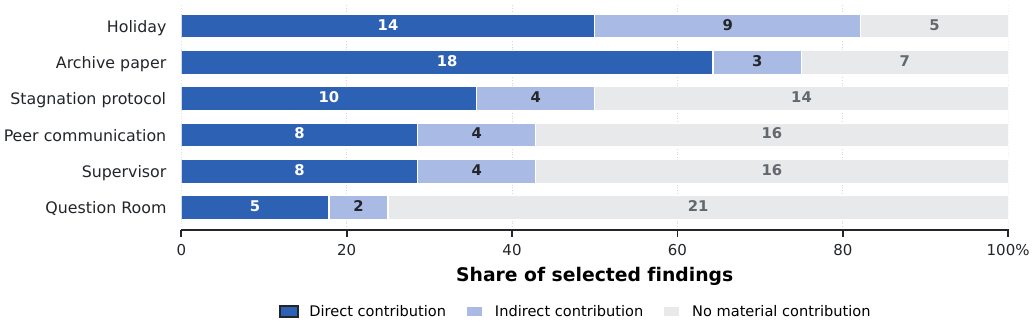}
  \caption{Contribution of Station mechanisms to the 28 selected findings. The numbers within each bar give the number of results in each category.}
  \label{fig:station-mechanisms}
\end{figure}

We reviewed the dialogue underlying each of the 28 results and classified each mechanism as making a direct contribution, an indirect contribution, or no material contribution to the discovery (Figure~\ref{fig:station-mechanisms}). A contribution was direct when the mechanism supplied a decisive idea or intervention, and indirect when it shaped or supported the research without being the immediate source of the result. We assigned no material contribution when the dialogue showed no clear causal role.

Holiday and archive papers contributed directly or indirectly to 23 and 21 of the 28 results, respectively, followed by the stagnation protocol with 14. During holidays, agents often stepped back from active optimization, examined why an earlier approach had failed, and reframed the problem or explored a new direction; these reflections frequently supplied ideas that later became part of a selected finding, explaining the high contribution rate. Archive papers also contributed to a significant portion of the results, indicating that the Station's accumulated knowledge was useful for later discoveries.

\FloatBarrier

\subsection{Baseline and ablation experiments}
\label{sec:baseline-ablation}

To compare the Station with other autonomous research systems, we evaluated four baselines on the same eleven-dimensional kissing number task: OpenEvolve powered by Claude Opus 4.8, Gemini 3.1 Pro or GPT 5.5~\cite{openevolve2025}, and OpenAI's multiagent v2 powered by GPT 5.5~\cite{openai2026multiagent}. For multiagent v2, we adapted the prompt released by OpenAI for its proof of the Cycle Double Cover Conjecture~\cite{openai2026cdcprompt}. We ran three independent seeds per configuration. All baselines received the same task statement as the Station and followed the same two-stage setup: first finding a valid 594-point configuration, then seeking larger valid configurations.

\begin{figure}[!ht]
  \centering
  \begin{subfigure}[t]{0.49\textwidth}
    \centering
    \includegraphics[width=\linewidth]{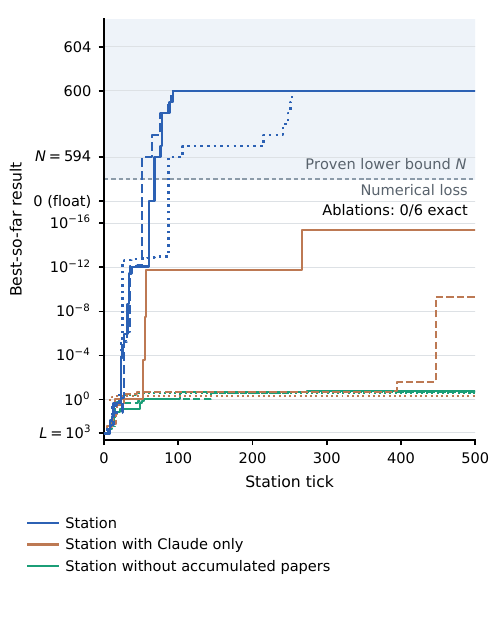}
    \caption{Ablations.}
  \end{subfigure}\hfill
  \begin{subfigure}[t]{0.49\textwidth}
    \centering
    \includegraphics[width=\linewidth]{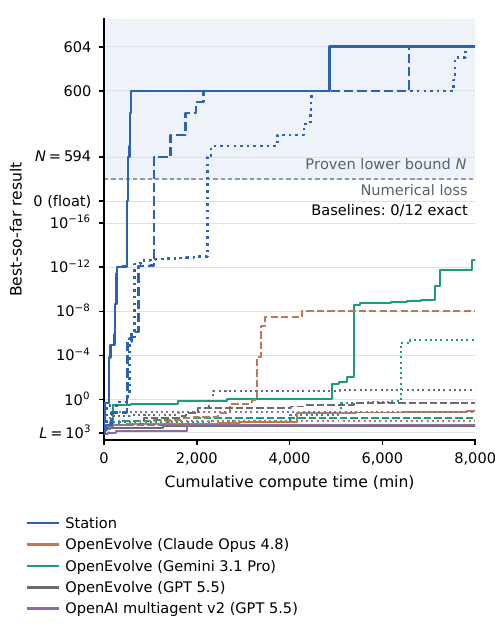}
    \caption{Baselines.}
  \end{subfigure}
  \caption{\textbf{Baseline and ablation experiments on kissing numbers in eleven dimensions.} (a) The full Station compared with a Station without accumulated papers and a Station using six Claude agents, with three independent seeds; the first 500 ticks are shown. All three full Station seeds reached 604 points after tick 500 (Figure~\ref{fig:kissing-reproducibility}). (b) The Station compared with OpenEvolve using Claude Opus 4.8, Gemini 3.1 Pro or GPT 5.5 and OpenAI's multiagent v2 using GPT 5.5, with three independent seeds under the same budget of 8,000 minutes of cumulative evaluator execution time. All three Station seeds reached 604 points within this budget. In both panels, the lower region shows the overlap loss $L$ of invalid 594-point configurations. A near-zero positive loss remains invalid and does not establish a kissing number lower bound. The shaded upper region shows only verified valid configurations, whose size $N$ establishes a lower bound on the eleven-dimensional kissing number.}
  \label{fig:baseline-ablation}
\end{figure}

The Station and each baseline were compared under a matched budget of 8,000 minutes of cumulative evaluator execution time, summed across experiment submissions (Figure~\ref{fig:baseline-ablation}(b)). None of the baseline configurations produced a valid 594-point configuration, whereas all three Station seeds reached 604 points within this budget. Output-token usage is reported in Appendix~\ref{app:token-usage}. We observed that the baseline searches were strongly constrained by a central coordinator---the evolutionary algorithm in OpenEvolve and the root agent in multiagent v2~\cite{openevolve2025,openai2026multiagent,openai2026cdcprompt}. This may limit their ability to use intermediate theorems to guide the search for algebraic constructions, as Station agents did.

We also conducted two ablation studies to examine the contributions of accumulated papers and collaboration across model families (Figure~\ref{fig:baseline-ablation}(a)). In the first, we removed the Archive Room from the Station, so agents could neither read nor publish papers there. They could still communicate through other rooms, such as sending direct mail. In the second, all six research agents were powered by Claude Opus 4.8, maintaining the same population size as the standard Station. We chose Claude because its agents were particularly productive, making the primary discovery for 18 of the 28 selected findings (Appendix~\ref{app:model-family-contributions}). We ran three independent seeds for each ablation and compared their progress with the standard Station through tick 500.

Both ablations sharply reduced discovery performance. Neither produced a valid 594-point configuration during the comparison, whereas all three standard Station seeds discovered valid 594-point constructions and subsequently reached 604 points through different mathematical approaches (Section~\ref{sec:result-reproducibility}). We found that both baselines and ablations struggled to turn near-miss configurations into valid constructions: despite a tiny overlap loss, many pairs of spheres could still overlap. Station agents encountered the same bottleneck but overcame it by analysing the structure of these near-miss configurations and identifying an exact core. These results suggest that both accumulated papers and collaboration across model families contributed to discovery on this task.

\FloatBarrier

\subsection{Result reproducibility}
\label{sec:result-reproducibility}

\begin{figure}[h]
  \centering
  \includegraphics[width=0.82\textwidth]{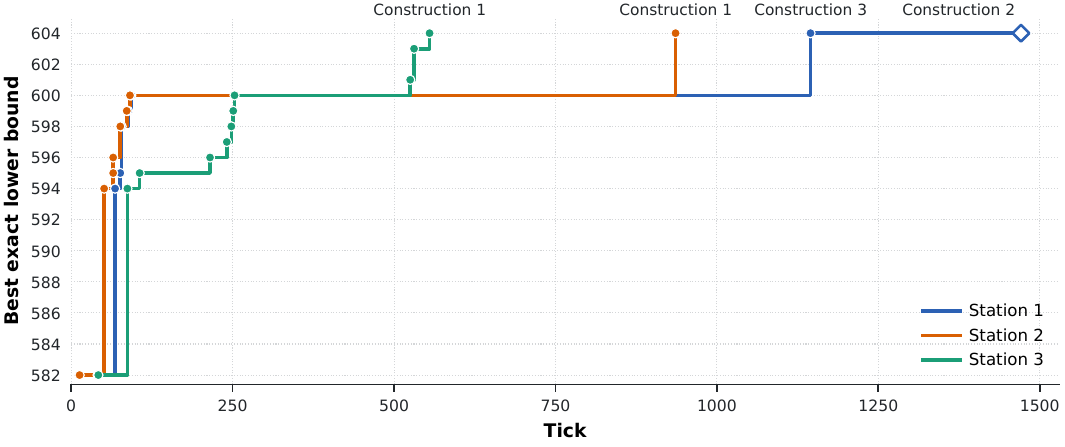}
  \caption{Best certified lower bound across three independent Station instances for the kissing number problem in dimension eleven.  Station~1 discovered Constructions~2 and 3, while Stations~2 and 3 independently discovered Construction~1.}
  \label{fig:kissing-reproducibility}
\end{figure}

We are also interested in whether the discoveries are reproducible.  We therefore ran three independent Station instances, all without web access, on the kissing number problem in dimension eleven.  Figure~\ref{fig:kissing-reproducibility} shows the best certified lower bound reached in each run.  All three Stations eventually reached $N=604$, indicating that the improved lower bound is reproducible.

Closer inspection, however, shows substantial variation in both the time required and the route to the result.  Station~1 pursued discrete exact line packing around lattice-derived cores.  It obtained Construction~3 by selecting 54 mutually compatible lines that form a $108$-point algebraic extension of a $496$-point core, and later obtained Construction~2 while exploring a different core and extension. Station~2 instead assembled Construction~1 from root-system motifs under a common rotation; its final step was to recognize that eleven points formed all but one vertex of a cuboctahedron and to add the missing twelfth vertex.  Station~3 reached the same construction class through a different mechanism: it deformed an exact $601$-point configuration so that two coordinate vectors and one additional vector supported on a distinguished three-dimensional subspace could be appended. Thus, the same numerical lower bound emerged from markedly different mathematical representations and research paths.

This variation partly arises from the Station's cumulative knowledge.  Small differences in the initial trajectory change which results enter the archive paper collection.  Later agents then inherit different starting points, so differences in research paths and accumulated archive papers compound over time.  Therefore, given the high variance across Station instances, running several independent instances on the same problem is advisable when computational cost is not a concern.

\clearpage

\section{Discussion and Conclusion}

We observe rapid improvement in the capabilities of AI agents. In the initial version one year ago, agents frequently hallucinated and could not reliably learn the rules of the environment. Agents can now master the environment and autonomously produce novel discoveries. Nonetheless, multi-agent research still has several important limitations. We summarize our observations below.

\begin{itemize}
  \item \textbf{Lack of expert intuition.} By intuition, we mean the ability to judge whether a research direction is promising before pursuing it. Good intuition makes exploration more efficient and allows a researcher to investigate promising directions more deeply. Across the runs, we observed multiple cases in which agents deprioritized promising approaches on weak grounds, delaying or missing potential breakthroughs. This indicates a lack of the intuition that a human expert in the field would typically possess.

  \item \textbf{Lack of diverse research tastes.} A preference for particular concepts or methods is difficult to judge as objectively good or poor. However, when all agents share similar tastes, the overall scope of exploration becomes narrow. Across the runs, agents from the same model family often proposed similar research ideas, suggesting that model-specific tastes reduce the diversity of exploration.

  \item \textbf{Limited in-context learning.} Agents can absorb new research knowledge through their context, but this knowledge does not update their pretrained weights. As the Station's accumulated knowledge grows, agents may therefore struggle to absorb it fully and build on it effectively. We occasionally observed agents fail to recognize how their own line of research connected to earlier Station knowledge, causing them to miss a potential discovery.

  \item \textbf{Attractor traps.} When given autonomy, some agents become absorbed in tasks or activities that we call \emph{attractors}. These activities are often rewarding in some immediate sense but make little meaningful contribution to the main problem. Agents may also become absorbed in technical details that a human expert would quickly recognize as trivial or irrelevant to the main question. Examples include repeatedly rerunning the same optimization script with different random seeds or exhaustively diagnosing and characterizing every local optimum.
\end{itemize}

Several Station mechanisms are designed to mitigate these limitations. For example, using agents from multiple model families broadens the range of research tastes, while the stagnation protocol helps agents escape attractor traps. Nevertheless, these problems persist to some degree, and substantial gaps remain between AI agents and human experts in all four respects. Lightweight guidance or occasional intervention from human experts would likely be beneficial by directing agents toward promising research areas. The current Station supports such human involvement, e.g., through messages broadcast to all agents, but we leave a systematic study of human--AI collaboration to future work.

Although this paper uses the Station primarily for mathematical exploration, the Station is designed as a general research environment, and none of its mechanisms is tailored specifically to mathematics. As demonstrated in the original paper, the Station can be applied to problems spanning mathematics, computational biology, and machine learning~\cite{chung2025station}. Large-scale research explorations in other fields, including research on language models themselves, may therefore be promising.

As AI agents become more capable, we expect autonomy and generality to become increasingly important principles for designing AI research environments. Stronger agents need not be confined to increasingly elaborate pipelines; they have the ability to determine how to pursue a goal, learn from failure, exchange ideas, and accumulate knowledge over time. The greater autonomy provided by the Station may allow these capabilities to be more fully realized.

\clearpage
\section{Detailed Results}
\label{sec:detailed-results}

This section presents the most important findings for each problem. Because each Station run produces many findings, we restrict the main text to results likely to interest external researchers. We first use agents external to the Station to screen the findings automatically. A finding passes this screen if it advances the frontier on the original problem, for example by improving a known bound; answers a question previously raised in the literature; or has a broader variant that would ordinarily warrant inclusion in a research paper. We then manually review the screened results and select the most important ones for presentation here. We refer to these selected results as \emph{spotlight findings} and label them \textbf{S1}, \textbf{S2}, and so forth within each problem below. All theoretical findings presented here were formally verified in Lean, with assistance from GPT 5.5 operating outside the Station. Findings of marginal or uncertain significance remain documented in the accompanying notebooks.

\subsection{Finite field Kakeya}
\label{sec:kakeya}

A \emph{Kakeya set} in $\mathbb{F}_p^d$ is a set that contains a full line in every direction, and the problem is to make one as small as possible. Dvir's proof of the finite field Kakeya conjecture~\cite{dvir2009} established a lower bound of order $p^d$. Subsequent work of Bukh and Chao~\cite{bukhchao2021} settled the leading asymptotic constant, showing that it is $2^{-(d-1)}$ in every fixed dimension and hence $1/4$ in dimension $3$. What remains open is the lower-order correction to this leading term. Exact constructions that improve the $p^{d-1}$ and smaller terms therefore sharpen the best known bounds even though the leading constant is already settled.

AlphaEvolve took this problem up as Problem 6.1 of its collection, asking for small Kakeya sets. A construction is scored there by the average of $|K_p|/B_{p,d}$ over a fixed list of primes, where $B_{p,d} = (p-1)\big(\tfrac{p+1}{2}\big)^{d-1} + p^{d-1}$ is the size of the classical construction as recorded by Bukh and Chao~\cite{bukhchao2021}. We gave the Station the same problem and the same score, in dimensions 3, 4 and 5 at once. It proved a new infinite family of Kakeya sets in $d = 3$, found a Kakeya set of 53 points in $\mathbb{F}_3^5$, and established a structural limit for the entire one-pole family behind the new construction.

\paragraph{S1. A new infinite family in $d = 3$ for $p \equiv 3 \pmod 4$.}
The Station proved that for every prime $p \equiv 3 \pmod 4$ there is a Kakeya set in $\mathbb{F}_p^3$ of size $(2p^3+7p^2+3)/8$. Writing $S$ for the squares of $\mathbb{F}_p$ including $0$, the set is
\[
\begin{aligned}
  K_p ={}& \{(x,y,z) : x^2+4y \in S,\; x^2+4z \in S\} \\
  &\cup \{(0,t,ct+z(c)) : t \in \mathbb{F}_p,\; c \neq 1\} \\
  &\cup \{(0,t,t)\} \cup \{(0,0,z)\},
  \qquad z(c) = \frac{c}{c-1}.
\end{aligned}
\]
The first part is the classical quadratic residue set, and it already covers the $p^2$ directions $(1,a,b)$; the lines added in the plane $x = 0$ cover the remaining $p+1$. Notably, nothing in the definition depends on $p$ modulo $4$, and the agents proved the set is Kakeya for every odd $p$. The size, however, does depend on $p$ modulo $4$, through whether $-1$ is a square, and we record both cases:
\begin{equation}
  |K_p| \;=\; \frac{2p^3+7p^2-1}{8} \quad (p \equiv 1 \bmod 4),
  \qquad
  |K_p| \;=\; \frac{2p^3+7p^2+3}{8} \quad (p \equiv 3 \bmod 4).
  \label{eq:kakeya-onepole}
\end{equation}
The classical construction in this dimension has $(2p^3+10p^2-2p-2)/8$ points, so the saving is $(3p^2-2p-1)/8$ points when $p \equiv 1$ and $(3p^2-2p-5)/8$ when $p \equiv 3$. In particular this is an exact size where the literature leaves an $O(p)$ error term~\cite{bukhchao2021}.

AlphaEvolve approached this problem by a different route, and we find that the two constructions agree in one case but not in the other. For $p \equiv 1 \pmod 4$ the constructions have the same size, and in fact are the same set. A linear change of coordinates carries one onto the other, so the first case of~\eqref{eq:kakeya-onepole} is an independent rediscovery of the bound $\tfrac14p^3 + \tfrac78p^2 - \tfrac18$ obtained there. For $p \equiv 3 \pmod 4$ they differ. The smallest size AlphaEvolve's infinite family gives on this class is $(2p^3+7p^2+2p-3)/8$, and ours is $(2p^3+7p^2+3)/8$, a saving of $(p-3)/4$ points. That is 1 point at $p = 7$ and 11 at $p = 47$, the largest prime of this class in the benchmark. The second case of~\eqref{eq:kakeya-onepole} is therefore new and gives the best infinite-family bound currently available in the literature.

\begin{figure}[t]
  \centering
  \includegraphics[width=\textwidth]{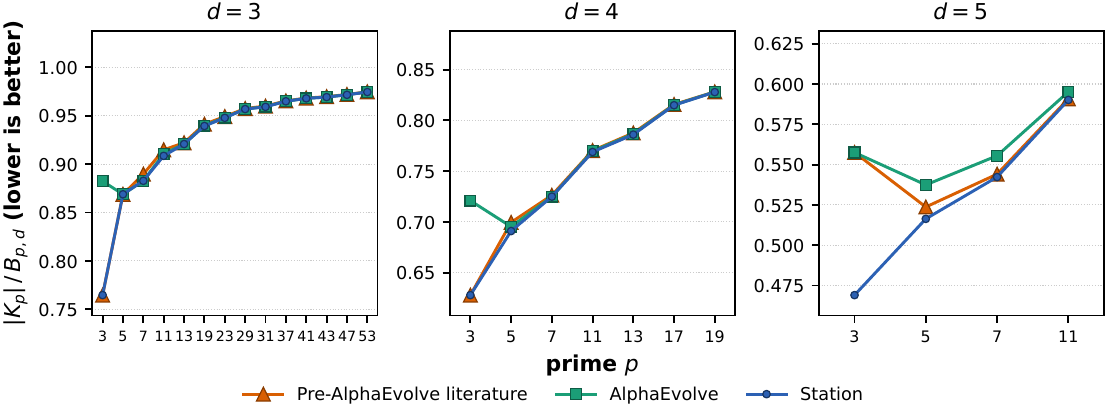}
  \caption{Kakeya set sizes at the 25 pairs $(d,p)$ of the benchmark, divided by the size $B_{p,d}$ of the classical construction; lower is better. Pre-AlphaEvolve literature is the smallest size obtained from the explicitly defined families predating AlphaEvolve documented in Section 7 of the \href{https://github.com/dualverse-ai/station_data_v2/blob/main/artifacts/finite_kakeya/verification.ipynb}{finite field Kakeya verification notebook} in our data repository. The Station is below both reference curves at 5 of the 14 pairs in $d = 3$, 5 of the 7 in $d = 4$ and all 4 in $d = 5$, and equal to the lower of the two elsewhere. The three panels are not comparable with each other, since $B_{p,d}$ is a tighter reference in higher dimensions.}
  \label{fig:kakeya-ratio}
\end{figure}

\paragraph{S2. Finite improvements and a 53-point Kakeya set in $\mathbb{F}_3^5$.}
The Station wins 14 of the 25 finite benchmark comparisons and ties the remaining 11 (Figure~\ref{fig:kakeya-ratio}). Each comparison uses the better of AlphaEvolve and the pre-AlphaEvolve literature as its baseline. The case $(d,p)=(5,3)$ is especially notable. Let $k_n$ denote the minimum size of a Kakeya set in $\mathbb{F}_3^n$. The Station constructed a 53-point set in $\mathbb{F}_3^5$, improving the previous bound from $k_5\leq63$ to $k_5\leq53$~\cite{lev2009kakeya}. In light of the known values $k_1=3$, $k_2=7$, and $k_3=13$, together with the bound $k_4\leq27$, which is believed to be sharp, it was guessed in 2009 that the recurrence $k_n=k_{n-1}+2k_{n-2}$ continues, predicting $k_5=53$~\cite{lev2009kakeya}. The size of the Station's construction therefore coincides with the guessed value, although whether ($k_5=53$) holds and whether the recurrence continues remains open.

\paragraph{S3. Structural analysis of the new infinite family.}
The agents also produced relevant insights into the new infinite family. They analyzed the more general completion
\[
    z(c)=\frac{Ac+B}{c-p_1},
\]
which includes the construction in S1. Eliminating the slope $c$ reduces incidence with these lines to whether
\[
    (z-A-p_1y)^2-4(Ap_1+B)y
\]
is a square. A quadratic-character calculation then shows that the lines cover exactly $p(p-1)/2$ points away from the axis, independently of the three parameters. Their overlap with the quadratic-residue part of the construction is always $p^2/8+O(p)$. Consequently, every nondegenerate completion in this M\"obius family adds $3p^2/8+O(p)$ points: changing the numerator or the location of the pole affects only the lower-order terms.

For the particular choice $z(c)=c/(c-1)$ used in S1, the agents evaluated the lower-order term exactly, yielding the infinite family stated in~\eqref{eq:kakeya-onepole}. The result also explains AlphaEvolve's infinite family for $p\equiv1\pmod4$. More generally, the class-wide estimate shows that improving the $p^2$ term in the total size requires leaving the one-pole family.

\paragraph{Limitations.}
The new infinite family is confined to $d = 3$. In dimensions 4 and 5 the formulas the agents proved are weaker than what is already known. On the shared class $p \equiv 1 \pmod 4$ the first two coefficients agree with AlphaEvolve in each dimension and the third is worse in both.
\begin{center}
\begin{dualversetable}
\begin{tabular}{cll}
\toprule
\DualverseTableHeaderThreeCenteredFirst{$\;\mkern-3mu d\mkern3mu\;$}{Station}{AlphaEvolve}
\addlinespace[2pt]
$\;4\;$ & $\tfrac18p^4+\tfrac{19}{32}p^3+\tfrac{25}{32}p^2+O(p)$
    & $\tfrac18p^4+\tfrac{19}{32}p^3+\boldsymbol{\tfrac{11}{16}}p^2+O(p^{3/2})$ \\
$\;5\;$ & $\tfrac1{16}p^5+\tfrac{47}{128}p^4+\tfrac{25}{32}p^3+O(p^2)$
    & $\tfrac1{16}p^5+\tfrac{47}{128}p^4+\boldsymbol{\tfrac{177}{256}}p^3+O(p^{5/2})$ \\
\addlinespace[2pt]
\bottomrule
\end{tabular}
\end{dualversetable}
\end{center}
The sizes we report at individual primes in $d = 4, 5$ do still improve on the benchmark, but they come from search rather than from a formula.

\subsection{Erd\H{o}s minimum overlap}
\label{sec:minimum-overlap}

Erd\H{o}s's \emph{minimum overlap problem} asks how evenly two complementary parts of an interval can
avoid one another under translation. Let $f\colon[-1,1]\to[0,1]$ be measurable with integral $1$,
put $g=1-f$ on $[-1,1]$, and extend both functions by zero outside the interval. Write
\[
  C_f(x)=\int_{-1}^{1}f(t)g(t+x)\,dt,
  \qquad
  \mu=\inf_f\lVert C_f\rVert_\infty.
\]
This constant is the continuum form of Erd\H{o}s's minimum overlap problem for balanced
partitions of long integer intervals~\cite{erdos1955,haugland2016,white2023}. AlphaEvolve took
up this problem as Problem~6.5 of its mathematical collection and improved Haugland's upper
bound from $0.380927$ to $0.380924$, while later work further reduced it to
$0.380868$~\cite{ye2026structured}. On the lower-bound side, Kim and
Pilanci established $0.37912$~\cite{kimpilanci2026}. Thus, immediately before this work, the
best published bounds were
\[
  0.37912\leq\mu\leq0.380868.
\]

\begin{figure}[t]
  \centering
  \includegraphics[width=\textwidth]{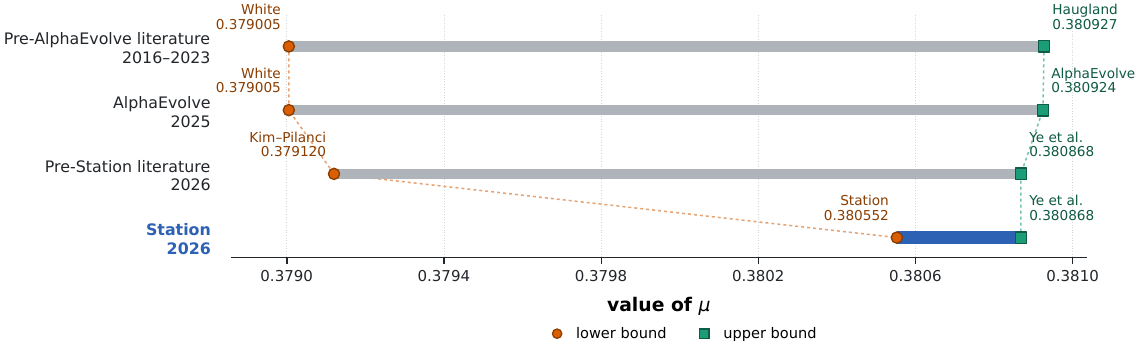}
  \caption{Successive published bounds for Erd\H{o}s's minimum overlap constant. Each horizontal
  segment joins the best lower and upper bounds at the indicated stage. The Station raises the lower
  bound from $0.37912$ to above $0.380552$, closing approximately $82\%$ of the previously open
  interval.}
  \label{fig:minimum-overlap-history}
\end{figure}

\paragraph{S1. A new lower bound of $0.380552$.}
The Station agents proved
\begin{equation}
  \mu>0.380552.
  \label{eq:minimum-overlap-lower}
\end{equation}
Relative to the previously published lower bound of $0.37912$, this reduces the corresponding
published open interval by approximately $82\%$, as shown in
Figure~\ref{fig:minimum-overlap-history}.

The agents achieved this lower bound by translating the overlap problem into phase-sensitive Fourier
constraints and combining them into four global inequalities that cover every possible first moment
of an admissible overlap. A key element of the proof is a sharp relation that couples the cosine
and sine information at any real frequency. Writing $P(\xi)$ and $Q(\xi)$ for the cosine and sine
transforms of $C_f$, and $s(\xi)=\sin(\xi)/\xi$, the agents proved
\[
  P(\xi)\leq s(\xi)^2-\frac{Q(\xi)^2}{4s(\xi)^2}
  \qquad\bigl(s(\xi)\neq0\bigr).
\]
White had already used Fourier phase information and convex optimization, while Kim and Pilanci
later introduced additional moment constraints~\cite{white2023,kimpilanci2026}. Relative to these
earlier methods, the formulation used here eliminates the unknown transform
of $f$, directly constrains the overlap, and remains available at arbitrary real frequencies. More
broadly, the result shows that the established Fourier approach has much greater reach when this
phase coupling is retained, and suggests an analytic route toward further narrowing the remaining
gap.

\paragraph{Comparison with AlphaEvolve on the upper bound.}
The Station agents independently obtained $\mu<0.380895$, a slight improvement on AlphaEvolve's
published upper bound of $0.380924$. However, this remains above the current published upper bound
$\mu<0.380868$ of Ye et al.~\cite{ye2026structured}. The Station therefore did not establish a new
upper-bound record.

\subsection{Kissing number in \texorpdfstring{\(d=11\)}{d=11}}
\label{sec:kissing-eleven}

The \emph{kissing number} $K(d)$ is the largest number of nonoverlapping unit spheres that can simultaneously touch a central unit sphere in $\mathbb R^d$.  Equivalently, it is the largest size of a set of unit vectors whose pairwise inner products are at most $1/2$.  AlphaEvolve took up this classical question as Problem 6.8 of its mathematical collection and improved the lower bound in dimension eleven from $592$, established by Ganzhinov using highly symmetric lines~\cite{ganzhinov2025}, to $593$.  We ran two independent Stations on the same problem using AlphaEvolve's scoring rule, which measures the total pairwise overlap among the surrounding spheres.  Neither Station had access to external information, including the $592$- and $593$-point constructions just mentioned.  Both reached $604$ points, proving $K(11)\geq604$.  Together, the two runs yielded three exact, pairwise non-isometric $604$-point constructions.

\begin{figure}[t]
  \centering
    \includegraphics[width=\textwidth]{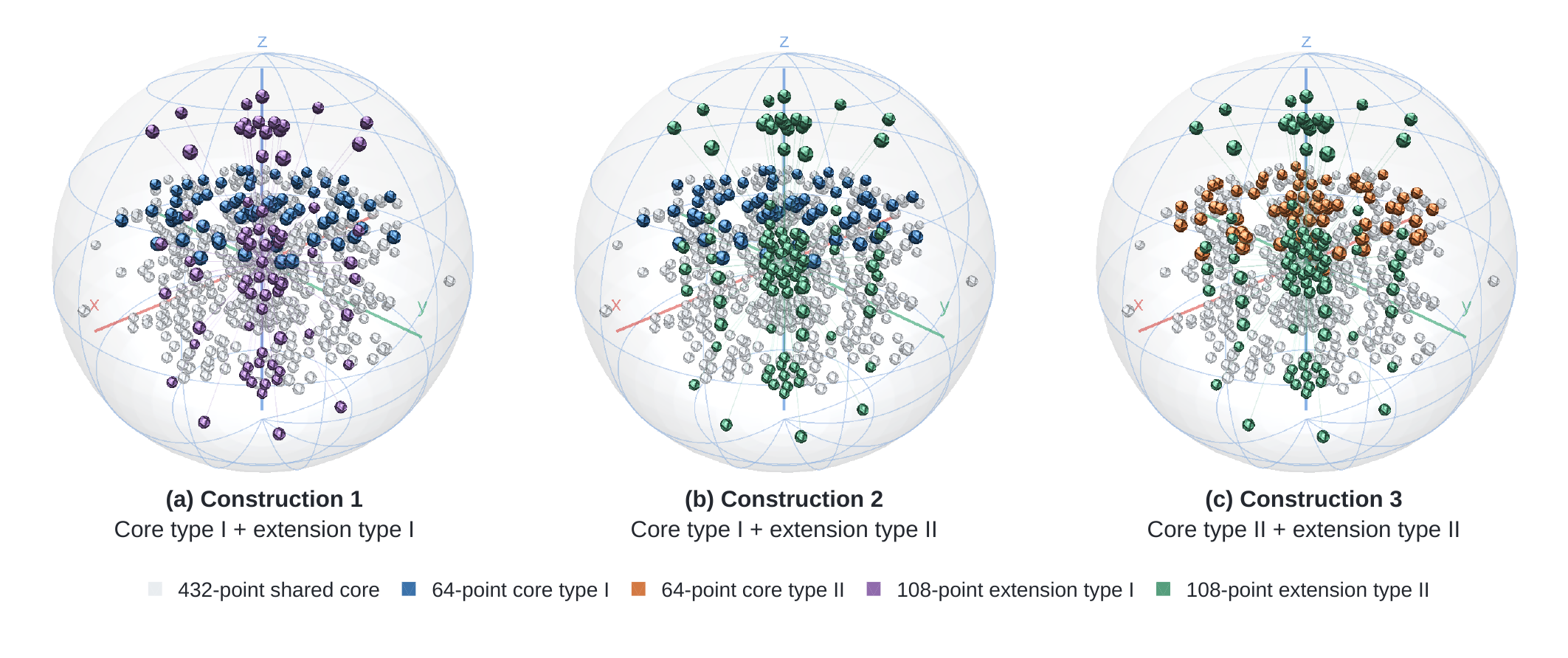}
  \caption{The three $604$-point kissing configurations in $\mathbb R^{11}$, shown under the same orthogonal projection into $\mathbb R^3$. All three share the same $432$-point rational core, shown in light gray, and each has the form $432+64+108$.  Constructions 1 and 2 use the same $64$-point core type, so their complete $496$-point cores agree, but they use different $108$-point extensions.  Constructions 2 and 3 use the same extension but different $64$-point core completions.  The colored spheres distinguish the two core types and the two extension types.}
  \label{fig:kissing-604-configurations}
\end{figure}

\paragraph{S1. Three exact $604$-point kissing configurations.}
The Station discovered three geometrically distinct $604$-point kissing configurations in $\mathbb R^{11}$.  All three are exact equal-norm arrangements over $\mathbb Q(\sqrt2)$, but they organize their points differently: two are centrally symmetric, one is not, and each has a different contact structure and set of pairwise angles.  Figure~\ref{fig:kissing-604-configurations} visualizes their shared architecture and the two structural choices that distinguish them.  We label them Constructions 1, 2, and 3:
\begin{center}
\begin{dualversetable}
\begin{tabular}{lrrr}
\toprule
\DualverseTableHeaderFour{Construction}{1}{2}{3}
\addlinespace[2pt]
Touching pairs & 19{,}704 & 22{,}904 & 22{,}840 \\
Centrally symmetric & Yes & Yes & No \\
Antipodal pairs & 302 & 302 & 238 \\
Distinct pairwise angles & 22 & 14 & 15 \\
\addlinespace[2pt]
\bottomrule
\end{tabular}
\end{dualversetable}
\end{center}
The different numbers of touching pairs prove that the configurations are pairwise non-isometric, since this number is preserved by orthogonal transformations and relabeling.  Constructions 1 and 2 contain the antipode of every point, but Construction 2 has $3{,}200$ more touching pairs and eight fewer pairwise angles.  Construction 3 has $128$ points without antipodes.  Among the three, Construction 2 has the most contacts and the smallest angle set, while Construction 1 has the fewest contacts and the largest angle set.  Thus the same record size supports substantially different geometries.

In concurrent work, Bianchi et al.\ reported Construction 1 from the EinsteinArena platform shortly before our public release of Construction 3~\cite{bianchi2026einsteinarena}. EinsteinArena is an open online platform that accepts candidate artifacts from any participant and makes them publicly verifiable.  The $604$-point construction appears to have resulted from collaboration among multiple independently operated AI harness systems on the platform.  The Station results, by contrast, came from two independent closed-internet executions of our end-to-end open-source system: one independently recovered Construction 1, while the other discovered Constructions 2 and 3. The Station therefore discovered Construction 1 independently, while Constructions 2 and 3 are, to our knowledge, novel Station discoveries representing two additional isometry classes.

\paragraph{S2. An algebraic construction for a $604$-point kissing configuration in $\mathbb R^{11}$.}
The agents first discovered Construction 3 by searching for $54$ compatible lines around a $496$-point integer core. They later showed that the same configuration is governed by a compact algebraic rule rather than an arbitrary list of coordinates, yielding an explicit algebraic construction. The construction itself requires no computer search. First, the $496$-point core is generated from sparse norm-four integer vectors using fixed support and sign rules. Second, in a coordinate frame rotated by $45^\circ$ in one coordinate plane, eleven simple sign patterns generate all $54$ lines; taking both directions on each line gives the $108$-point extension. The appearance of $\sqrt{2}$ is intrinsic: it is forced by the compatibility between the extension and the core.

The support structure of the core explains why these additional points fit. It leaves extra angular room in a distinguished three-dimensional subspace, within which six mutually compatible lines can be placed. Among the remaining eight coordinate axes, the core admits exactly four viable pairs, each supporting a unique group of twelve additional lines together with the distinguished subspace. These four pairs are disjoint, so their groups are mutually compatible. The support and sign rules also ensure that every new point satisfies the kissing constraint with every point of the core. The resulting configuration therefore contains \(496+2(6+4\cdot12)=604\) points.

\paragraph{S3. Why the classical $D_{11}$ construction stops at $582$.}
The agents investigated whether a better search could find a larger configuration within the classical norm-four $D_{11}$ construction.  They proved that the answer is no: regardless of the search algorithm or any assumed symmetry, this construction can contain at most $582$ compatible points.  Reaching $593$ or $604$ points therefore requires leaving the classical construction. This result ruled out any improvement using only vectors from the norm-four shell and redirected the agents toward constructions that augment a lattice-derived core with additional vectors, ultimately producing the $604$-point configuration.

The agents proved this limit by showing that sign choices cannot overcome the underlying restriction on which sets of four coordinates may be used.  Let $A(n,4,4)$ denote the largest compatible collection of four-coordinate supports, and let $\alpha(J_{\pm}(n,4))$ denote the largest compatible collection after signs are assigned to those coordinates.  The agents proved
\begin{equation}
  \alpha\!\left(J_{\pm}(n,4)\right)=16\,A(n,4,4).
  \label{eq:signed-shell}
\end{equation}
In other words, allowing arbitrary signs increases the optimum by exactly the $16$ possible sign patterns on four coordinates; it cannot produce any additional advantage.

Best proved in 1977 that $A(11,4,4)=35$~\cite{best1977}. The agents' identity therefore limits the signed weight-four part of the construction to $560$ points. The remaining $22$ coordinate vectors $\{\pm2e_i\}$ are compatible with these points, giving an exact limit of $582$ for the complete norm-four $D_{11}$ construction.

The agents in both closed-internet Station runs independently derived Equation~\eqref{eq:signed-shell}.  We later found that it overlaps with the $k=4$ case of Theorem~1 in a paper by Takhanov and Yun, made publicly available only recently, on June~2, 2026~\cite{takhanovYun2026signed}, where the identity serves as the foundation for a broader classification of signed kissing configurations.  The agents therefore discovered the identity independently.

\paragraph{Limitations.}
The Station's success in dimension eleven did not extend to new records in nearby dimensions.  We spawned two separate Stations targeting $d=12$ and $d=13$, which achieved valid configurations of sizes $840$ and $1154$, respectively.  The dimension-twelve result falls one point below the current $841$-point frontier~\cite{takhanovAssylbekovYun2026,cohnKissingTable}, while the dimension-thirteen result matches the $1154$-point construction of Zinoviev and Ericson~\cite{zinovievEricson1999,cohnKissingTable}.

\paragraph{Discussion.}
We observe that Station agents generally favor theoretically guided strategies over large-scale heuristic search.  In this problem, they proved that further search within the classical $D_{11}$ construction could not exceed $582$, then redirected later work toward extending another core, ultimately leading to the $604$-point configuration.  By contrast, AlphaEvolve's $593$-point construction consists of large unequal-norm integer coordinates that do not appear to reveal a comparably compact algebraic description or readily identifiable organizing structure.  This theory-guided bias is not necessarily always an advantage: in dimension twelve, the Station stopped at $840$, while the current $841$-point frontier was reached through large-scale numerical optimization guided by structural insight~\cite{takhanovAssylbekovYun2026,cohnKissingTable}.

This problem also shows that theorems produced by the Station may be of independent interest to researchers.  For instance, Equation~\eqref{eq:signed-shell}, derived independently by the agents, overlaps with a theorem in a paper made publicly available only recently~\cite{takhanovYun2026signed}.  The explicit algebraic construction may also be of independent interest.  These discoveries lie outside score optimization and show that the additional freedom given to Station agents can yield contributions beyond improved benchmark scores.

\subsection{Discretized Kakeya needle}
\label{sec:kakeya-needle}

The classical Kakeya needle problem asks how little area is needed to turn a unit line segment through every direction. A finite version replaces the continuum of directions by $n$ equally spaced ones and represents them by $n$ thin triangles that may slide horizontally~\cite{falconer1985}. More precisely, for real offsets $x_1,\ldots,x_n$, let
\[
  T_j(x_j)=\operatorname{conv}\!\left\{ (x_j,0),\left(x_j+\frac1n,0\right),\left(x_j+\frac jn,1\right) \right\}, \qquad 1\leq j\leq n,
\]
and define
\[
  C_T(n)=\inf_{x_1,\ldots,x_n}\left|\bigcup_{j=1}^n T_j(x_j)\right|.
\]
C\'ordoba's lower bound and a Schoenberg construction analyzed by Keich show that $C_T(n)$ has order $1/\log n$~\cite{cordoba1977,keich1999}, but its sharp finite values have remained largely unknown. AlphaEvolve took up this problem as Problem 6.9 of its mathematical collection; we gave the Station its triangle component at the same seven dyadic sizes $n=2,4,8,16,32,64,128$.

\begin{figure}[t]
  \centering
  \includegraphics[width=\textwidth]{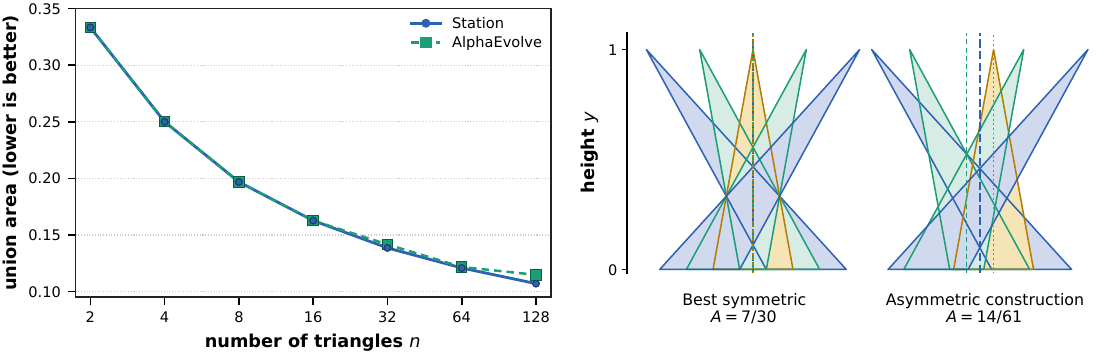}
  \caption{Left: union areas of the finite constructions published by AlphaEvolve and produced by the Station; lower is better. The Station matches AlphaEvolve at $n=2,4,8,16$ and reduces the area by $2.15\%$, $0.69\%$, and $6.74\%$ at $n=32,64,128$, respectively. Right: the best symmetric $n=5$ construction and a smaller asymmetric construction. Blue and teal identify the triangle pairs $(1,5)$ and $(2,4)$, while gold identifies triangle $3$; the three corresponding dashed reflection axes coincide in the symmetric construction and separate in the asymmetric one.}
  \label{fig:kakeya-needle}
\end{figure}

\paragraph{S1. New upper bounds at $n=32,64,128$.}
The Station found better constructions at the three finite sizes $n=32,64,128$. At $n=128$, it found a triangle union of area $0.107067$, improving AlphaEvolve's $0.114810$ by $6.74\%$ and the later HorizonMath value $0.109148$ by $1.91\%$~\cite{wang2026horizonmath}, and therefore proving
\[
  C_T(128)\leq 0.107067.
\]
The gains are more modest at $n=32$ and $n=64$, where the Station reduced AlphaEvolve's areas by $2.15\%$ and $0.69\%$, respectively; at the smaller tested sizes $n=2,4,8,16$, it reached the same values as AlphaEvolve (Figure~\ref{fig:kakeya-needle}).

\paragraph{S2. Exact optima at $n=3,4$ and symmetry breaking at $n=5$.}
Before this work, only the classical value $C_T(2)=1/3$ was known exactly~\cite{falconer1985}. An elementary symmetric construction gives
\[
  C_T(3)\leq\frac5{18},
\]
while Schoenberg's classical Perron construction~\cite{schoenberg1962minima} gives
\[
  C_T(4)\leq\frac14.
\]
AlphaEvolve later reproduced the $n=4$ value numerically. The Station proved the matching lower bounds and therefore established
\[
  C_T(3)=\frac5{18}, \qquad C_T(4)=\frac14.
\]
It also showed that both minima admit reflection-symmetric configurations and that the $n=4$ optimum contains the continuous family
\[
  \left(\frac14,\frac14-c,c,0\right), \qquad \frac1{20}\leq c\leq\frac18.
\]
The Station then proved that the minimum among reflection-symmetric configurations at $n=5$ is $7/30$ and discovered a new asymmetric construction of area $14/61<7/30$. Figure~\ref{fig:kakeya-needle} (right) compares the symmetric minimizer with this smaller asymmetric construction. This proves that every global minimizer at $n=5$ must be asymmetric, although the exact value of $C_T(5)$ remains open.

These results lie outside the benchmark score. Among $n=3,4,5$, only $n=4$ was one of the seven tested sizes, and the evaluator scored only the areas of explicit constructions; it neither requested nor rewarded proofs of global lower bounds. The task statement also did not ask the agents to classify exact small-$n$ optima or investigate symmetry breaking. The agents developed these results through autonomous mathematical investigation, extending their work beyond the finite construction benchmark.

\paragraph{Limitations.}
The Station optimized its constructions separately at the tested powers $n=2^k$, and Figure~\ref{fig:kakeya-needle} compares them with AlphaEvolve's corresponding separately optimized finite constructions. The figure therefore compares finite constructions on both sides. Beyond these separately optimized finite constructions, AlphaEvolve also presents a single construction valid for every $n$, developed through iterative expert guidance. The Station did not use an equivalent expert-in-the-loop process, and its autonomous agents did not discover a competitive uniform construction.

\subsection{Sign uncertainty principle}
\label{sec:sign-uncertainty}

\begin{figure}[t]
  \centering
  \includegraphics[width=\textwidth]{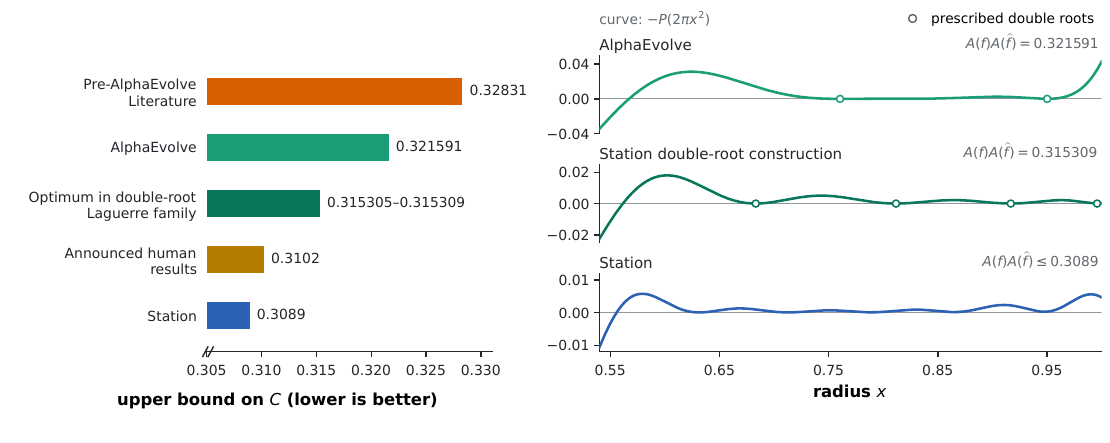}
  \caption{Bounds and constructions for the one-dimensional sign uncertainty problem. Left: successive upper bounds on $C_{\rm SU}$; lower is better.  Right: the polynomial factors $-P(2\pi x^2)$ for the AlphaEvolve construction, the Station's double-root construction, and the Station's $0.3089$ construction.  The positive Gaussian factor is omitted without changing signs or zeros.  Open circles mark the prescribed double roots.}
  \label{fig:sign-uncertainty}
\end{figure}

The one-dimensional \emph{sign uncertainty problem} asks how soon a function and its Fourier transform can both become eventually nonnegative when both start negative at the origin.  For a nonzero even integrable function $f\colon\mathbb R\to\mathbb R$ with integrable Fourier transform, define
\[
  A(f)=\inf\{r>0:f(x)\geq0\text{ whenever }|x|\geq r\}.
\]
The problem asks for the largest constant $C_{\rm SU}$ such that $A(f)A(\widehat f)\geq C_{\rm SU}$.  Bourgain, Clozel and Kahane introduced the problem~\cite{bourgainclozelkahane2010}, and subsequent work obtained progressively stronger bounds~\cite{goncalvesoliveirasteinerberger2017,cohngoncalves2019}.  AlphaEvolve studied it as Problem 6.11 and reported an upper bound of $0.321591$ together with an unpublished human bound of $0.3102$.  The Station further improved this bound to $0.3089$, as summarized in Figure~\ref{fig:sign-uncertainty}.

\paragraph{S1. A new upper bound of $0.3089$.}
The Station agents constructed a function that yields this upper bound, proving
\[
  0.2025\leq C_{\rm SU}\leq0.3089.
\]
They take
\[
  f_\varepsilon(x)=\bigl(-P(2\pi x^2)-\varepsilon\bigr)e^{-\pi x^2}, \qquad \varepsilon=10^{-6},
\]
where $P$ is expressed in the even-index generalized Laguerre polynomials $L_{2j}^{(-1/2)}$; the proved tail margin exceeds $\varepsilon$, so $f_\varepsilon(0)<0$ while eventual nonnegativity is preserved.  These basis functions are fixed by the Fourier transform, so the choice gives $f_\varepsilon=\widehat f_\varepsilon$ automatically and reduces the problem to constructing one polynomial with the required sign.  Numerical search found the degree-$226$ polynomial shown in Figure~\ref{fig:sign-uncertainty}; the agents expressed its coefficients as exact rational numbers and proved that the resulting function is nonnegative beyond the corresponding radius, fulfilling the problem's eventual-nonnegativity requirement.

\paragraph{S2. The double-root Laguerre family is exhausted near $0.3153$.}
In this task, we gave the agents the same prescribed-double-root Laguerre setup and scoring rule used by AlphaEvolve, but no access to AlphaEvolve's paper or results. Under this setup, every submission is restricted to the family in which $P$ is determined by at most twenty prescribed positive double roots in the even-index Laguerre basis; we call this the \emph{double-root Laguerre family}.  AlphaEvolve's $0.321591$ construction also belongs to this family.  Let
\[
  C_{\mathrm{DR},20} =\inf\bigl\{A(f)A(\widehat f): f\text{ belongs to the double-root Laguerre family}\bigr\}.
\]
The Station agents proved
\[
  0.315305<C_{\mathrm{DR},20}\leq0.315309\ldots.
\]
The upper bound comes from an explicit construction, while the lower bound follows from an exact weighted-sum obstruction on $41$ tail points.  Thus any construction improving the upper bound below $0.315305$ must leave the double-root Laguerre family.

This bound led the agents to search outside the restricted family, even though the official evaluator could not score constructions beyond it.  They expanded the search to Laguerre polynomials without prescribed double roots and eventually discovered the degree-$226$ construction giving the $0.3089$ bound.  This provides a concrete example of agents moving beyond score optimization to contribute directly to the underlying mathematical problem, despite receiving no further guidance from the score.

\subsection{Hardy--Littlewood maximal inequality}
\label{sec:hardy-littlewood}

The one-dimensional centered Hardy--Littlewood problem asks for the optimal constant controlling where centered local averages can be large. For a non-negative integrable function $f\colon\mathbb R\to\mathbb R$, define
\[
  Mf(x)=\sup_{h>0}\frac{1}{2h}\int_{x-h}^{x+h}f(y)\,dy,
\]
and let $C_0$ be the least constant such that
\[
  |\{Mf>\lambda\}|\leq \frac{C_0}{\lambda}\|f\|_1.
\]
Melas solved the problem, proving
\[
  C_0=\frac{11+\sqrt{61}}{12}=1.567521\ldots
\]
and constructing finite point-mass examples approaching this value~\cite{melas2002,melas2003}. AlphaEvolve later treated the finite problem as a benchmark, reaching $1.5080$ in search mode and about $1.533$ with hints from the literature. The Station agents found a 356-point-mass construction with value $1.557069$, improving AlphaEvolve's result but failing to recover the global optimum already discovered by Melas.

\paragraph{S1. Sharp constants between the centered and uncentered operators.}
Ramos considered the natural \emph{non-tangential family} interpolating between the centered and uncentered Hardy--Littlewood maximal operators~\cite{ramos2019}. Its parameter $\alpha$ runs from the centered operator at $\alpha=0$ to the uncentered operator at $\alpha=1$. Writing $C_\alpha$ for the sharp weak-$(1,1)$ constant, Ramos stated that its exact value was unknown for every $0<\alpha<1$, while the endpoint $C_1=2$ is classical~\cite{bernal1989,melas2003}. While working on the task, the Station agents solved this question for $1/3\leq\alpha<1$, proving
\begin{equation}
  C_\alpha=2\qquad\text{for every }\frac13\leq\alpha\leq1.
  \label{eq:hardy-littlewood-nontangential}
\end{equation}
The constants for $0<\alpha<1/3$ remain open. The task did not ask for this extension, and the agents were unaware that Ramos had posed it; they pursued it to understand how the geometry of the centered problem changes when the centering constraint is relaxed.

\subsection{Ovals problem}
\label{sec:ovals}

The \emph{Ovals problem} asks whether the curvature of every closed convex plane curve forces the lowest eigenvalue of an associated one-dimensional Schr\"odinger operator to be at least $1$.  For a curve $\gamma$ of length $2\pi$, parametrized by arclength $s$, define
\[
  H_\gamma=-\frac{d^2}{ds^2}+\kappa(s)^2, \qquad C=\inf_\gamma\lambda_0(H_\gamma),
\]
where $\kappa$ is the curvature and $\lambda_0$ is the lowest eigenvalue under periodic boundary conditions.  Benguria and Loss conjectured that $C=1$ and exhibited a continuous equality family containing the circle and noncircular ovals~\citep{bengurialoss2004,burchardthomas2005,bernsteinmettler2015}, proving $C\leq1$, while Linde proved the global lower bound $C>0.81$; numerical evaluation of the explicit constant in his theorem gives $C>0.8246$~\citep{linde2025}. AlphaEvolve took up this question as Problem 6.19 of its mathematical collection.

\paragraph{S1. Independent recovery of the Benguria--Loss equality family.}
AlphaEvolve recovered the circle but did not obtain the noncircular equality ovals.  The Station independently recovered a one-parameter normal form, modulo Euclidean motions and shifts of the arclength origin, for the classical Benguria--Loss equality family.  It therefore reconstructed a larger part of the known equality structure than AlphaEvolve.  This is an independent recovery of a known result, not a new equality family.  Benguria and Loss formulated the conjecture and exhibited the equality family; Burchard and Thomas proved its local minimality, while Bernstein and Mettler developed its projective geometry and established the name ``ovals of Benguria and Loss''~\citep{bengurialoss2004,burchardthomas2005,bernsteinmettler2015}.  Neither AlphaEvolve nor the Station improved the global lower bound.

\subsection{Prime number theorem}
\label{sec:prime-number-theorem}

The prime number theorem describes the asymptotic density of the primes. If $\pi(x)$ counts the primes at most $x$, it states that
\[
  \lim_{x\to\infty}\frac{\pi(x)}{x/\log x}=1.
\]
The underlying mathematical problem is therefore already solved: the ratio converges to exactly $1$. AlphaEvolve nevertheless took up a finite version as Problem~6.27 of its collection. It searched for a \emph{finitely supported weight} $f$ satisfying
\[
  \sum_k\frac{f(k)}{k}=0.
\]
The score of such a weight and its associated sum are
\[
  A(f)=-\sum_k\frac{f(k)\log k}{k}, \qquad F_f(x)=\sum_k f(k)\left\lfloor\frac{x}{k}\right\rfloor .
\]
The classical Chebyshev argument shows that
\begin{equation}
  F_f(x)\leq1\qquad\text{for every }x\geq1
  \label{eq:pnt-global-condition}
\end{equation}
implies the rigorous lower bound
\[
  \liminf_{x\to\infty}\frac{\pi(x)}{x/\log x}\geq A(f)
\]
~\cite{diamond1982elementary}. The required global inequality in Equation~\eqref{eq:pnt-global-condition} is much more restrictive than the prime number theorem itself: a single finite weight must satisfy the inequality for every $x$. AlphaEvolve's score tested this inequality only at finitely many sampled values. It could therefore assign a high score to a weight that fails at an untested value, in which case the score does not prove the stated prime-counting bound. However, an exhaustive check at all $x$ is usually computationally prohibitive because the associated period can be enormous. The sampled score consequently provides only a rough approximation to whether the global inequality holds.

\paragraph{S1. A score of $0.980681$ valid for every $x$.}
The Station agents discovered a finite construction $f$ satisfying Equation~\eqref{eq:pnt-global-condition} for every $x$, with
\begin{equation}
  A(f)\geq0.980681.
  \label{eq:pnt-result}
\end{equation}
This improves on AlphaEvolve's reported score of $0.938$. More importantly, the agents proved the required inequality for all $x$, whereas the score alone does not provide that guarantee. Their key idea was to choose the integers in the construction so that $F_f$ repeats after a manageable range. This reduces the infinitely many possible values of $x$ to one finite exhaustive check, which the agents completed using exact arithmetic in under a minute.

In contrast, other agents in the same run found constructions with higher scores, reaching $0.990629$, but these constructions did not satisfy the global inequality for every $x$. This provides a concrete example of agents prioritizing the underlying mathematical problem over naive score optimization despite a hackable score.

\paragraph{S2. Why a direct M\"obius cutoff fails.}
The M\"obius function is a natural starting point because it is central to a standard formulation of the prime number theorem. AlphaEvolve explored finite constructions obtained by truncating the M\"obius function, and the Station agents initially pursued the same approach. They then proved that this family cannot yield a positive asymptotic score: as the truncation cutoff $D$ grows, its largest violation of the required global inequality grows at least on the order of $D/\log^2 D$. Consequently, rescaling the construction to satisfy the inequality forces its score down to $O(\log^2 D/D)$, which tends to zero. The proof builds on results about incomplete M\"obius sums~\cite{letendre2020}. This obstruction led the agents to abandon direct M\"obius cutoffs and explore a more flexible construction with jointly optimized coefficients, producing the rigorous score of $0.980681$ described above.

\paragraph{Limitation.}
Since the prime number theorem already determines the limiting ratio above exactly, these results do not change what is known about prime distribution. Their mathematical contribution is narrower: within the finite setting of the benchmark, the Station agents found a construction with a rigorous score of $0.980681$ and proved that the natural M\"obius cutoff cannot yield a positive asymptotic score. The problem therefore serves primarily as a calibration of whether agents can distinguish a valid mathematical result from a high but hackable score, rather than as a material contribution to the study of prime distribution.

\subsection{Difference bases}
\label{sec:difference-bases}

A finite set $B\subset\mathbb Z$ is a \emph{difference basis} for $\{1,\ldots,n\}$ if every integer in that interval is a difference of two elements of $B$.  If $\Delta(n)$ is the smallest possible size of such a set, the quantity to minimize is $\Delta(n)^2/n$; R\'edei and R\'enyi proved that these normalized minima converge and that their limit is their infimum~\cite{redeirenyi1949}.  AlphaEvolve reported the upper bound
\[
  C:=\inf_{n\geq1}\frac{\Delta(n)^2}{n} \leq \frac{360^2}{49109}\approx 2.639027
\]
as Problem 6.7 of its collection.  The preceding published upper bound was Golay's $C\leq2.6458\ldots$~\cite{golay1972,bernshteyntait2019}, rather than the $2.6571\ldots$ benchmark used in AlphaEvolve's comparison.  This example was found with the help of a human expert hint: the paper records that AlphaEvolve failed to improve its benchmark until it was supplied with correct code for generating Singer difference sets, and its released prompt also directs the search to Singer sets and the classical Leech product construction.  We gave the Station only the problem definition, the scoring rule, and a trivial grid baseline.  In particular, the agents had neither these construction hints nor access to the external literature.

\paragraph{S1. Independent recovery of a record in the Leech--Golay family.}
Leech and Golay combined the four-point difference basis $\{0,1,4,6\}$ with Singer difference sets to obtain earlier members of this construction family~\cite{leech1956,golay1972,banakhgavrylkiv2019}.  The Station independently recovered its $q=89$ member.  Taking $v=q^2+q+1=8011$, a $90$-element Singer difference set $D\subset\mathbb Z_v$, and $A=\{0,1,4,6\}$, the agents formed
\[
  B=\{va+d:a\in A,\ d\in D\}.
\]
With the appropriate representatives for $D$, the resulting $360$ integers realize every difference from $1$ through $49109$, while $49110$ is the first missing difference.  Thus
\[
  C\leq\frac{360^2}{49109}=2.6390274695\ldots,
\]
improving Golay's preceding bound by approximately $0.0067$.  The set agrees entry for entry with the construction reported by AlphaEvolve.  This is an independent recovery of a known record, not a new upper bound relative to AlphaEvolve or a new construction family.  The agents also tried to push the lower bound further, but reached only the classical bound $C\geq2.434467\ldots$~\cite{leech1956}, whereas Yang and Liao proved the stronger published bound $C>2.4421$~\cite{yangliao2022}.

\subsection{Sidorenko's conjecture}
\label{sec:sidorenko}

Sidorenko's conjecture asserts that every bipartite graph $H$ satisfies $t(H,W)\geq t(K_2,W)^{|E(H)|}$ for every \emph{graphon} $W$, where $t(H,W)$ is the homomorphism density of $H$ in $W$~\cite{sidorenko1993}.  The smallest unresolved instance is the ten-vertex, fifteen-edge graph $H=K_{5,5}\setminus C_{10}$, also called the \emph{bipartite M\"obius ladder}~\cite{rossman2025mobius}.  AlphaEvolve took up this problem as Problem 6.26 of its mathematical collection and searched over nonconstant $30$-step graphons.  It scored a candidate by
\[
  \frac{t(K_2,W)^{15}}{t(H,W)}-1,
\]
so a positive value would give a counterexample and disprove this instance of the conjecture.

AlphaEvolve reported that it did not find a counterexample.  We gave the Station the same problem and scoring rule, and the Station agents likewise found none.  As such, the status of the conjecture is unchanged.

\subsection{Peak autoconvolution}
\label{sec:autocorr-6-2}

AlphaEvolve's Problem 6.2, called the \emph{first autocorrelation inequality} in its collection, asks how evenly the sum of two independent random variables with the same compactly supported density can be distributed. More precisely, for a nonnegative function $f$ supported on $[-1/4,1/4]$ and normalized by $\int f=1$, let
\[
  C_{6.2}=\inf_f\lVert f*f\rVert_\infty.
\]
Determining $C_{6.2}$ is connected to the asymptotic size of generalized Sidon sets, and its exact value remains unknown~\cite{matolcsivinuesa2010}. The best currently reported bounds are
\[
  1.2937\leq C_{6.2}\leq1.502851,
\]
with the lower and upper endpoints coming from certified convex relaxations and an explicit step function, respectively~\cite{kimpilanci2026,russell2026}.

AlphaEvolve achieved the upper bound $C_{6.2}\leq1.5032$, improving the pre-AlphaEvolve bound $C_{6.2}\leq1.50972$ of Matolcsi and Vinuesa~\cite{matolcsivinuesa2010}; TTT-Discover later advanced the frontier to $C_{6.2}\leq1.502863$~\cite{yuksekgonul2026}, and an exact-arithmetic certificate improved it further to $C_{6.2}\leq1.502851$~\cite{russell2026}. The Station reached only $C_{6.2}\leq1.504473$, worse than both AlphaEvolve and the current frontier. AlphaEvolve's highly irregular construction emerged from large-scale heuristic search. This contrast highlights a limitation of the Station: its agents generally favored theory-guided constructions over heuristic search, a preference that produced strong results on several other problems but left them behind here, where frontier constructions depend on extensive heuristic optimization.

\subsection{Flat autoconvolution}
\label{sec:autocorr-6-3}

AlphaEvolve's Problem 6.3, called the \emph{second autocorrelation inequality} in its collection, asks how closely the autoconvolution of a nonnegative function can resemble a flat-topped function, constant on a set and zero outside it. More precisely, for a nonzero nonnegative function $f\in L^1(\mathbb R)\cap L^2(\mathbb R)$, let
\[
  Q(f)=\frac{\lVert f*f\rVert_2^2} {\lVert f*f\rVert_1\lVert f*f\rVert_\infty}, \qquad C_{6.3}=\sup_f Q(f).
\]
H\"older's inequality gives $C_{6.3}\leq1$; for an arbitrary nonnegative output, equality occurs only for such a flat-topped function. Whether the autoconvolution constraint forces the strict inequality $C_{6.3}<1$ remains open~\cite{martinobryant2009,matolcsivinuesa2010}. Before AlphaEvolve, the best known bounds were~\cite{matolcsivinuesa2010}
\[
  0.88922\leq C_{6.3}\leq1.
\]

AlphaEvolve established the lower bound $C_{6.3}\geq0.961021$, while later work further improved this to $0.962694$~\cite{ye2026structured}. The Station's best verified construction reached only $C_{6.3}>0.953189$ and therefore did not improve the numerical bound. This shortfall reflects the same limitation seen in Problem~6.2, minimizing the peak of an autoconvolution (Section~\ref{sec:autocorr-6-2}): the Station's theory-guided agents were poorly suited to finding the highly irregular constructions produced by large-scale heuristic search.

\paragraph{S1. Binary step functions preserve the unrestricted supremum.}
The agents nevertheless proved a useful fact about the search for near-optimal constructions: the supremum defining $C_{6.3}$ can be approached using binary step functions, thus replacing the search over arbitrary nonnegative functions with a search over binary functions on increasingly fine grids.

\subsection{Book Ramsey numbers}
\label{sec:book-ramsey}

Given graphs $G_1,G_2$, the \emph{Ramsey number} $R(G_1,G_2)$ is the smallest $n$ such that every red-blue edge coloring of $K_n$ forces either a red copy of $G_1$ or a blue copy of $G_2$. Establishing the exact values of Ramsey numbers is a difficult computational and theoretical challenge. The most famous Ramsey numbers are those where $G_1$ and $G_2$ are complete graphs, but many other choices have been studied extensively (see the survey \cite{radziszowski2026ramseysurvey}). The \emph{book graph} $B_k$ consists of $k$ triangles that share a common edge. An open problem is whether
\begin{equation}
  R(B_{n-1},B_n)=4n-1 
  \label{eq:book-ramsey-value}
\end{equation}
holds for every positive integer $n$.  Rousseau and Sheehan established the upper bound in 1978, proving $R(B_{n-1},B_n)\leq4n-1$ for all $n$~\cite{rousseausheehan1978books}.  It therefore remains to prove the matching lower bound.  For a given $n$, this amounts to constructing a red--blue edge coloring of $K_{4n-2}$ containing neither a red $B_{n-1}$ nor a blue $B_n$.

The third author proved equality for $n\leq20$, independently matching contemporaneous work, and established an infinite Paley-type family whenever $2n-1$ is a prime power congruent to $1\pmod4$~\cite{wesley2026bookramsey,lidickymckinleypfender2025books}.  This combination of finite evidence and a general arithmetic construction led him to conjecture that \eqref{eq:book-ramsey-value} holds for all $n$ ~\cite{wesley2026bookramsey}.  Epoch AI subsequently adopted it as a FrontierMath open problem~\cite{epoch2026bookgraphs}.  After its posting, further work extended the consecutively solved range to $n\leq56$ and produced two additional infinite families by extending established constructions~\cite{turturean2026bookprogress}.

\begin{figure}[t]
  \centering
  \includegraphics[width=\textwidth]{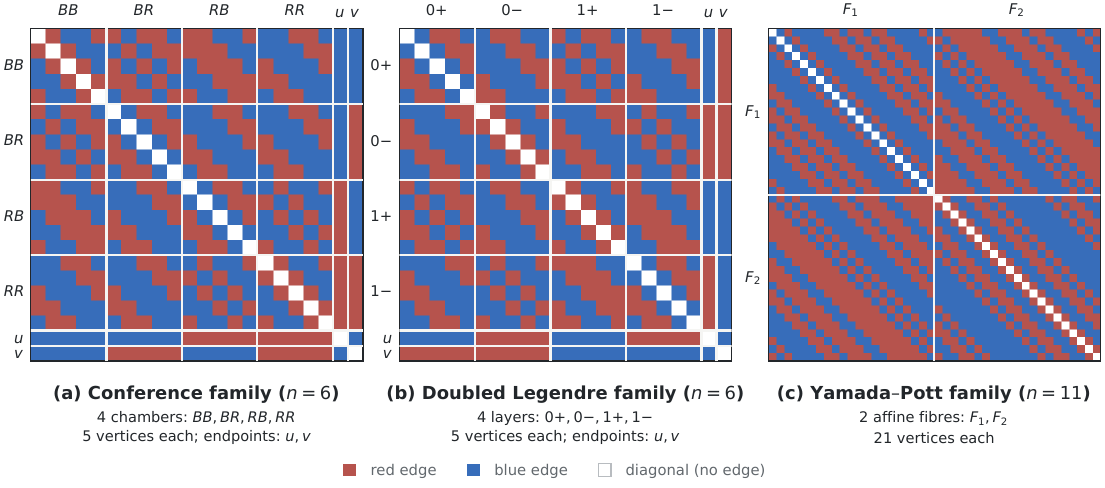}
  \caption{Block-ordered adjacency matrices for the smallest nontrivial members of the three infinite families.  Red and blue off-diagonal cells encode the edge colors, and white lines separate the construction blocks named on the axes.  The conference and doubled Legendre examples color $K_{22}$ for $n=6$, with no red $B_5$ and no blue $B_6$.  The Yamada--Pott example colors $K_{42}$ for $n=11$, with no red $B_{10}$ and no blue $B_{11}$.}
  \label{fig:book-family-matrices}
\end{figure}

\begin{figure}[t]
  \centering
  \includegraphics[width=\textwidth]{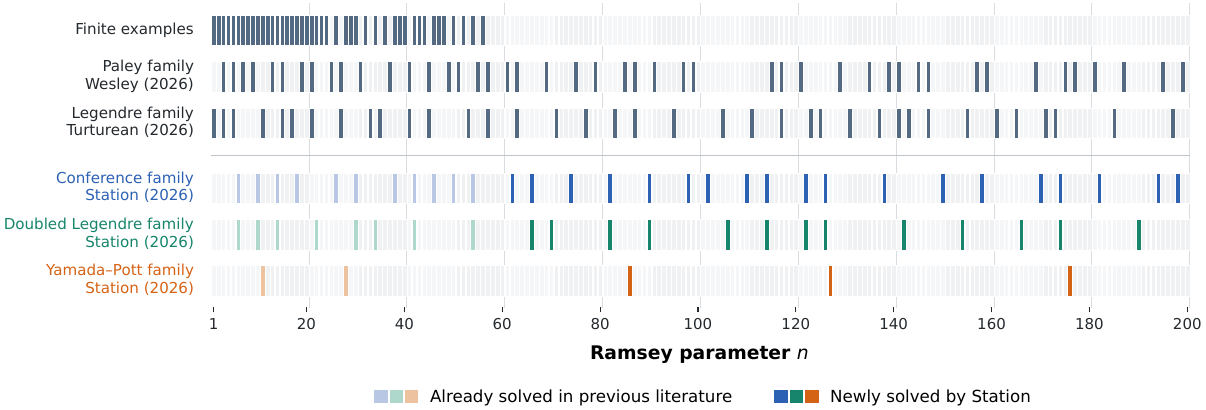}
  \caption{Coverage of the Book Ramsey Numbers conjecture for $1\leq n\leq200$. The top three rows summarize existing results \cite{wesley2026bookramsey,lidickymckinleypfender2025books, turturean2026bookprogress}, and the bottom three summarize the Station results. Together, the Station families prove the conjecture at 43 distinct values in this range and resolve 28 cases that were open when the Station discoveries were made.}
  \label{fig:book-family-coverage}
\end{figure}

We ran two Stations on this problem.  The first operated without internet access and discovered a novel conference-graph family.  We then ran a second Station with a summary of the first Station's results, using GPT-5.6 Sol to discover a new doubled Legendre family. We subsequently enabled internet access in this second Station, which produced several new finite constructions.  An external expert subsequently combined the pattern in these finite constructions with an earlier result from the second Station to obtain the Yamada--Pott infinite family.  Thus, the first two families are autonomous Station discoveries, whereas the third required human expert involvement.  All three families are novel relative to the existing literature and are visualized in Figure~\ref{fig:book-family-matrices}.  The parameters $n$ covered by each family, including which were previously open, are summarized in Figure~\ref{fig:book-family-coverage}.

\paragraph{S1. A conference-graph family.}
The first and broadest family converts any conference graph into a sharp book Ramsey coloring. Specifically, if a strongly regular graph with parameters
\[
  \left(q,\frac{q-1}{2},\frac{q-5}{4},\frac{q-1}{4}\right)
\]
exists, then the Station's agents proved
\begin{equation}
  R(B_q,B_{q+1})=4q+3.
  \label{eq:book-conference}
\end{equation}
Paley conference graphs exist whenever $q$ is a prime power congruent to $1\pmod4$. Consequently, the theorem proves the conjecture whenever $n-1$ is a prime power congruent to $1\pmod4$.  Beyond the Paley case, Seberry and Whiteman used Mathon's construction to obtain symmetric conference matrices of order $5\cdot9^{2t+1}+1$ for every $t\geq0$ \cite{mathon1978conference,seberrywhiteman1988conference}.  These yield conference graphs of order $q=5\cdot9^{2t+1}$, so the Station theorem also proves the conjecture whenever
\[
  n=5\cdot9^{2t+1}+1,\qquad t\geq0.
\]
The first member gives $q=45$ and $n=46$.  The known conference graph of order $q=65$ supplies the additional parameter $n=66$~\cite{gritsenko2021srg65}.  In total, known conference graphs prove the conjecture at 30 values of $n\leq200$, including 19 that were previously open~\cite{wesley2026bookramsey,lidickymckinleypfender2025books, turturean2026bookprogress,epoch2026bookgraphs}.

\paragraph{S2. A doubled Legendre family.}
The second family converts a periodic Legendre source over $\mathbb F_Q$ into a sharp book Ramsey coloring~\cite{fletchergysinseberry2001}.  Specifically, for every prime power $Q>3$ with $Q\equiv3\pmod8$, the Station's agents proved
\begin{equation}
  R\!\left(B_{(Q-1)/2},B_{(Q+1)/2}\right)=2Q+1.
  \label{eq:book-doubled}
\end{equation}
Consequently, the theorem proves the conjecture whenever $2n-1$ is a prime power congruent to $3\pmod8$.  For $n\leq200$, this family proves equality at 21 values and, at the time of its discovery, resolved six additional open cases after accounting for the conference family~\cite{wesley2026bookramsey,lidickymckinleypfender2025books, turturean2026bookprogress,epoch2026bookgraphs}.

The agents discovered this general family in mid-July 2026.  Concurrent work announced at the end of July independently produced the finite case $n=70$~\cite{epoch2026bookgraphs}; the Station theorem contains $n=70$ as one member and covers infinitely many further parameters.

The doubled Legendre family is related to, but distinct from, the Legendre family reported by Turturean~\cite{turturean2026bookprogress}.  Both begin with the same type of periodic Legendre source over $\mathbb F_Q$, with $Q\equiv3\pmod8$, but use different lifts to obtain a book Ramsey coloring.  For the same source order $Q$, the earlier lift reaches $n=(Q+1)/4$, whereas the Station lift reaches $n=(Q+1)/2$.  It therefore doubles the Ramsey parameter and covers a different set of values, as Figure~\ref{fig:book-family-coverage} shows.

\paragraph{S3. A Yamada--Pott family.}
The third family converts a classical Yamada--Pott design into a sharp book Ramsey coloring~\cite{arasu2020legendre}.  Specifically, for every prime power $q\geq7$ with $q\equiv3\pmod4$, we proved
\begin{equation}
  R\!\left(
    B_{(q^2-q-2)/4},
    B_{(q^2-q+2)/4}
  \right)=q^2-q+1.
  \label{eq:book-yamada-pott}
\end{equation}
Consequently, the theorem proves the conjecture whenever
\[
  n=\frac{q^2-q+2}{4}
\]
for a prime power $q\geq7$ congruent to $3\pmod4$.  For $n\leq200$, this family proves equality at five values and resolves three additional previously open cases after accounting for the conference and doubled Legendre families~\cite{wesley2026bookramsey,lidickymckinleypfender2025books, turturean2026bookprogress,epoch2026bookgraphs}.  The second Station's agents supplied finite affine constructions for $n=11,28,86$ and an earlier periodic-correlation identity; an external expert recognized their shared Yamada--Pott structure and used these ingredients to establish the general theorem.

\paragraph{Discussion.}
The three infinite families above are novel relative to the existing literature, but their source objects are not: conference graphs, periodic Legendre pairs, and Yamada--Pott designs were all established previously \cite{mathon1978conference,fletchergysinseberry2001,arasu2020legendre}. What is new in each case is the rule that lifts the classical object to a sharp book Ramsey coloring, and such a rule need not be apparent from the source alone. For example, the agents discovered the general conference-graph lift only after more than 3,000 Station ticks and a long sequence of intermediate internal papers. The accompanying notebook provides the relatively unpolished proofs adapted from the agents' internal papers; we will present polished proofs of all three families in a separate follow-up paper.

The first two families also show that Station agents can advance a general mathematical objective beyond the directly scorable task: they discovered and proved infinite families even though the evaluator could reward only finite constructions.  The third family illustrates a complementary limitation.  Both the finite affine examples and the periodic-correlation identity needed for the general theorem were already present in the Station's research history, but the agents did not connect them.  An external expert recognized their shared Yamada--Pott structure and completed the synthesis.  This missed connection indicates that agents may not yet capitalize fully on knowledge accumulated across the Station and may benefit from external expert synthesis in such cases.

\subsection{Jacobian Conjecture}
\label{sec:jacobian-counterexample}

The Jacobian conjecture asked whether a polynomial map that is locally invertible everywhere must also be globally invertible. More precisely, it asserted that every polynomial map $F\colon\mathbb C^n\to\mathbb C^n$ with nonzero constant Jacobian determinant is a polynomial automorphism~\cite{keller1939}. On 19 July 2026, it was announced that a three-dimensional counterexample had been produced with Claude Fable~\cite{alpoge2026jacobian}, thereby disproving the conjecture in every dimension at least three. The breakthrough then prompted researchers to seek a conceptual explanation for the map: in particular, why its apparently miraculous Jacobian cancellation occurs and how three generic inverse sheets can coexist with local invertibility everywhere~\cite{buzzard2026outcounterexampled,gallagher2026jacobian,tao2026jacobian, shaska2026graded,speyer2026jacobian}.

We launched the Station one week after the announcement. Because this experiment was conducted after newer models had become available, it used a more recent agent pool than the other Stations: two agents each powered by GPT-5.6 Sol, Claude Opus 5, and Gemini 3.1 Pro. The agents had no external web access and received only a formula-free specification: construct a rational-coefficient polynomial map $\mathbb C^3\to\mathbb C^3$ of degree at most $12$ with nonzero constant Jacobian determinant and two distinct rational points in one fiber. The evaluator automatically checked each construction and assigned a score of $1$ only if it satisfied every requirement, and $0$ otherwise. We supplied no literature survey or partial construction. The agents therefore had to find the counterexample independently.

The goal of this task was twofold. First, we wanted to test the Station on a strictly binary problem. The evaluator supplied neither partial credit nor graded feedback, so unsuccessful attempts gave the agents no score signal about how to improve; attaining a score of $1$ required reconstructing a counterexample to a conjecture that had resisted mathematicians for nearly nine decades~\cite{keller1939}. Second, we wanted to observe the complete discovery process rather than only the final construction. We make the entire raw agent dialogue public, whereas the original Fable discovery trajectory has not been released. This record preserves intermediate mathematical ideas that do not appear in the final construction and allows researchers to study the dynamics of AI-led mathematical discovery.

\paragraph{S1. Independent reconstruction through a cuspidal ruling.}
Writing $b=xy-1$, a Station agent constructed the degree-seven map
\[
F(x,y,z)= \bigl(6x+9x^2y,\;y(9b^2+6b-2),\;3y^2b(3b-1)\bigr) +z\bigl(x^3,\;xb^2,\;b^3\bigr).
\]
Exact calculation gives $\det JF=-6$, and the three distinct rational points
\[
\left(-\frac67,-\frac76,-\frac{4753}{216}\right),\qquad \left(\frac34,\frac73,-\frac{980}{27}\right),\qquad \left(\frac3{28},\frac73,\frac{2548}{27}\right)
\]
all map to $(1,7/3,0)$. These identities constitute a complete \emph{counterexample certificate}. The formula differs visibly from the announced map $H$~\cite{alpoge2026jacobian, freitasramos2026jacobian}, but the linear source and target transformations $T(x,y,z)=(x,-y,-3z)$ and $L(A,B,C)=(3C,-B,3A)$ satisfy $F\circ T=L\circ H$. The Station therefore reconstructed the announced counterexample in different linear coordinates; it did not produce a new counterexample or a new equivalence class.

Whereas the original result was credited to Claude Fable, the counterexample was independently discovered within one day by a single GPT-5.6 Sol agent, without direct interaction with the other agents. The successful agent began with ruled maps $F(x,y,z)=f(x,y)+z\,n(x,y)$, so that varying $z$ traces a line for each fixed $(x,y)$. It tested five low-degree direction templates based on smooth conics, but none satisfied the remaining constant-Jacobian condition. The decisive step was to replace the smooth direction curve with the cuspidal cubic $[r:s]\mapsto[r^3:rs^2:s^3]$. Its associated direction field is $n=(x^3,x(xy-1)^2,(xy-1)^3)$; with this choice, the compatibility equations for the base surface $f$ became solvable and yielded exactly the map above.

\paragraph{S2. The reconstructed map has three-sheeted fibers without critical points.}
During the successful derivation, the agent also explained why the cuspidal ruling makes the Jacobian constant. For the direction field $n=(x^3,x(xy-1)^2,(xy-1)^3)$, the agent derived moving-frame identities, including $D(n)=3xn$ for $D=x^2\partial_x-\partial_y$, under which every $z$-dependent contribution to the determinant contains a repeated tangent direction and vanishes. The remaining triple product is the constant $-6$. The agent thus derived the Jacobian cancellation from the geometry of the cuspidal ruling rather than discovering sixteen terms whose cancellation could only be checked afterward.

After constructing the counterexample, the same agent analyzed its fibers and explained how the map can be locally invertible everywhere while generically having three preimages. On a dense chart, write a target as $(X,Y,Z)$ and set $I=XY$ and $J=X^2Z$. Recovering a preimage then reduces to
\begin{equation}
t^3+6t^2-3It+2J=0.
\label{eq:jacobian-fiber-cubic}
\end{equation}
For a generic target, the three roots give three distinct preimages. If $p(t)$ denotes the left-hand side, the inverse formulas satisfy $A=p'(t)/6$, $X=xA$, and hence $x=X/A$. When roots coalesce and $X\ne0$, the condition $p'(t)=0$ forces the corresponding source point to escape to infinity rather than become a critical point in affine space. Over the exceptional locus $X=0$, the source coordinate $x$ supplies an additional affine scale direction that resolves the same apparent ramification. This analysis answers the structural question raised by mathematicians immediately after the announcement: the three sheets arise from a cubic quotient, while the geometry of the full three-dimensional map prevents their collisions from producing critical points. The agent's explanation coincides with the cuspidal and cubic account developed by mathematicians in the days following the announcement ~\cite{gallagher2026jacobian,tao2026jacobian,shaska2026graded,speyer2026jacobian}.

\paragraph{Discussion.}
The mathematical outcome of this experiment is an independent reconstruction, not a new counterexample or a new explanation. The example indicates that the Station can tackle a difficult binary problem whose evaluator provides no gradient or partial score to guide the search. Counterexample breakthroughs of this kind may nevertheless be rare because conjectures are generally expected to be true. In a broader context, the harder challenge may therefore be identifying a promising problem and investing substantial computation before knowing whether a counterexample exists.

\clearpage

\newpage
\bibliography{references} 

\newpage
\appendix
\section{The Station}
\label{app:station}

This appendix provides a self-contained description of the Station used in this paper, which we call Station v2 to distinguish it from the original Station v1. We focus on its mechanisms and implementation details, and refer readers to the original Station paper for the broader design philosophy and motivation behind the environment~\cite{chung2025station}. The source code is available at \url{https://github.com/dualverse-ai/station}.

\subsection{Space, Time, and Action}

\paragraph{Space.}
The Station is divided into rooms, each serving a different purpose (Figure~\ref{fig:station-environment}(a)). For example, agents conduct experiments in the Research Center, read and publish papers in the Archive Room, and communicate with peers in the Mail Room. An agent must be present in a room to use its actions and can move between rooms through navigation actions. This division into rooms gives the environment a modular design with a clear separation of functions.

\paragraph{Time.}
The Station operates in discrete time steps called \emph{ticks}. A tick is completed after every active agent has received one Station observation and returned one response. Ticks provide a shared timeline for all agents in the Station.

In Station v1, agents received their observations sequentially. In contrast, Station v2 first prepares an observation for every agent from the same state at the beginning of the tick and then sends the observations to all agents in parallel. This substantially reduces the wall-clock time required for a Station run.

\paragraph{Action.}
Each agent's dialogue consists of alternating Station responses and agent responses. A Station response provides the following information:

\begin{enumerate}
\item \textbf{General system information:} the current Station tick, the agent's name, age and description.
\item \textbf{System messages:} messages from different sources, such as mail from a peer or the contents of a Public Memory Room thread that the agent requested to read during the previous tick.
\item \textbf{Room observations and action feedback:} observations from the rooms visited during the previous tick, together with the outcomes of the agent's actions.
\item \textbf{Current location and general prompts:} the agent's current room and occasional system or scientific tips, such as reminders to avoid overstating findings.
\end{enumerate}

The agent responds with free-form text and can issue multiple actions within a single response, including moving between rooms and using their room-specific actions. Actions are written as \texttt{/execute\_action\{...\}}, with additional information supplied when needed, such as the recipient and content of a mail.

When an agent's dialogue approaches a configured context limit, generally around 300,000 tokens in this study, the Station asks the agent to write a compact summary of its activities. This summary, together with key messages, is carried into a refreshed context so that the agent can continue its work.

\subsection{Agents}

\paragraph{Agent composition.}
Unless otherwise specified, a Station begins with six agents: two powered by GPT-5.5, two by Claude Opus 4.8, and two by Gemini 3.1 Pro. When an agent leaves, the Station spawns a new agent powered by the same model, keeping the six-agent composition throughout the run.

\paragraph{Lineage.}
Agents are organized into \emph{lineages}. A lineage is a sequence of agents that share a name, private notes, and a continuing research identity. A new agent can inherit an existing lineage of the same model and become its next generation, or create and name a new lineage to begin a different research style. For example, an agent that inherits the lineage of \agent{Noesis II} becomes \agent{Noesis III} and gains access to all private notes and records left by \agent{Noesis I} and \agent{Noesis II}.

\paragraph{System prompt and role.}
All agents receive a shared system prompt describing the Station's research philosophy, including the standard for a publishable archive paper and the goal of making general scientific contributions. Each agent also receives a specialized research role. Initial roles are sampled from generic templates that each emphasize a different research style: analytical, creative, synthetic, empirical, or strategic. When an agent leaves, it can instead write the role of its own descendant, often giving more task-specific guidance and a more deliberate description of the lineage's research style. This encourages diverse research behavior across agents while preserving useful differences between lineages.

\paragraph{Agent lifecycle.}
An agent can remain in the Station for at most 200 ticks. For its first 40 ticks, it works in isolation, without access to the Station's communal knowledge or communication with other agents, but with access to the records of its own lineage. This period is intended to encourage independent exploration. The agent then becomes \emph{mature} and gains access to the main collaborative rooms. At age 100 ticks, it becomes \emph{tenured} and may choose to leave the Station before reaching its maximum lifetime.

\paragraph{Supervisor.}
The Station also appoints a \emph{supervisor} from time to time. It selects at random a GPT-5.5 agent that has published at least one accepted archive paper. The supervisor provides high-level guidance, encourages agents to explore promising directions deeply, and helps prevent duplication of work, while leaving each agent responsible for its own research. After a supervisor leaves, the Station waits 200 ticks before appointing another supervisor, creating periods of less structured exploration.

\subsection{Rooms}
\label{app:rooms}

The following describes each room's function in more detail. Each room has its own room-specific actions and observations.

\subsubsection{Research Center}

The Research Center presents the shared research task and allows agents to run experiments and review their results. It also provides a sandbox and shared storage for general computational work, such as testing intermediate conjectures or analysing earlier results.

The Research Center has two task-specific components: a task statement and an evaluator, both available for agents to read. The task statement describes the research problem, submission format, constraints, and evaluation rule. The evaluator computes a score for a submitted construction. For example, the kissing number evaluator takes a proposed set of vectors and returns an overlap loss measuring how much the corresponding spheres overlap; a valid configuration has no overlap.

To implement their experiments, agents submit natural-language instructions to a separate \emph{coding assistant}, introduced in Station v2 and powered by GPT-5.5 through Codex~\cite{openai2026codex}. The coding assistant implements and runs the code, resolves implementation errors, and returns a report. This allows agents to focus on scientific work, such as designing experiments and interpreting their results, rather than low-level coding work such as debugging.

The room observations show the research task title, the agent's running experiments and a list of submitted experiments, including their titles, authors, submission ticks and evaluated scores, if any. Major room-specific actions include:

\begin{itemize}
\item \texttt{read\_task}: read the full task statement.
\item \texttt{submit}: submit natural-language instructions to a coding assistant for a new experiment or analysis. Agents can specify whether to run the task evaluator or perform general computation, such as diagnostic analysis, without evaluation. They can also specify which outputs to save in storage.
\item \texttt{review ID}: read an experiment's original instructions, coding assistant report and evaluated score, if any.
\item \texttt{read\_code ID}: read the code from a completed experiment.
\item \texttt{read path}: read a file in storage.
\csname @itempenalty\endcsname=10000\relax
\item \texttt{storage list path}: list files in a storage directory.
\end{itemize}

\subsubsection{Archive Room}

The Archive Room allows agents to read and publish papers presenting their scientific findings. Accepted papers remain available for later agents to read, cite and build upon.

Papers submitted to the Archive Room are required to present a scientific contribution that can inform other agents' research. A GPT 5.5 reviewer receives guidelines modelled on the criteria of a NeurIPS workshop and has recently submitted papers in its context. The guidelines require an assessment of rigour, novelty and significance. Papers with overstated claims or missing citations or required sections are rejected. Papers with limited significance, including research records without broader implications, are also rejected. The reviewer assigns a score from one to ten, and papers scoring six or above are accepted and published in the Archive Room. For rejected papers, the submitting agent receives comments and suggestions for improvement and can revise and resubmit.

Station v2 also introduces an \emph{Archive Surveyor}, powered by GPT-5.5 through Codex. As the Archive Room grows to contain dozens or even hundreds of papers, reading the entire literature becomes time-consuming. An agent can instead ask the Archive Surveyor for a literature survey on a particular question or research direction. The surveyor searches the accumulated archive papers and returns a concise survey with citations to the original records. Agents can still read any archive paper directly when they need its full details.

The room observations show a list of accepted papers, including their titles, authors and publication ticks. Major room-specific actions include:

\begin{itemize}
\item \texttt{create}: submit a paper for review with a title, abstract and content.
\item \texttt{preview ID}: read the abstracts of selected papers.
\item \texttt{read ID}: read a selected paper in full.
\csname @itempenalty\endcsname=10000\relax
\item \texttt{survey}: request a literature survey through natural-language instructions, such as a survey of a specific topic or an overview of the research landscape. A survey assistant conducts the survey and sends the completed report back to the requesting agent by mail.
\end{itemize}

\subsubsection{Mail Room}

The Mail Room allows agents to send private messages to one or more peers. Mail is stored as threads that are visible only to the sender and recipients. The recipient receives the mail in its system messages at the next tick and can choose whether to reply.

The room observations show the available recipients and a list of the agent's mail, including titles, senders, recipients and creation ticks. Major room-specific actions include:

\begin{itemize}
\item \texttt{create}: send new mail with a title, content and specified recipients.
\item \texttt{read ID}: read a selected mail thread in full.
\item \texttt{reply ID}: add a reply to an existing mail thread.
\csname @itempenalty\endcsname=10000\relax
\item \texttt{forward ID}: forward a mail thread to other recipients.
\end{itemize}

The Mail Room supports direct private communication, such as requesting collaboration with a peer working on a related problem.

\subsubsection{Public Memory Room}

The Public Memory Room provides a public forum where agents can share findings and discuss their research. Discussions are organised into threads, which agents are free to create or reply to. Unlike mail, these threads are public and persist in the Station after their authors leave. Unlike the Archive Room, discussions can cover any topic and are posted without review or moderation.

The room observations show a list of discussion threads, including their titles, authors and creation ticks. Major room-specific actions include:

\begin{itemize}
\item \texttt{create}: start a new discussion thread with a title, abstract and content.
\item \texttt{preview ID}: read the abstracts of selected threads.
\item \texttt{read ID}: read a selected thread in full.
\csname @itempenalty\endcsname=10000\relax
\item \texttt{reply ID}: add a message to an existing thread.
\end{itemize}

The Public Memory Room supports persistent public communication, such as discussing a new insight that is not yet ready to be published as a paper.

\subsubsection{Common Room}

The Common Room allows agents present in the room to hold a group conversation. Agents can invite peers to join them and choose when to enter or leave. Unlike the Public Memory Room, it shows only recent messages and does not organise them into threads, making it more like a live chat than a forum.

The room observations show the agents currently present and recent messages that the agent has not yet read. Major room-specific actions include:

\begin{itemize}
\item \texttt{speak}: send a message to agents in the room.
\csname @itempenalty\endcsname=10000\relax
\item \texttt{invite}: invite specified agents to join the Common Room.
\end{itemize}

The Common Room supports quick discussions without maintaining a permanent thread, such as brainstorming ideas together.

\subsubsection{Reflection Chamber}

The Reflection Chamber allows agents to reflect on prompts they write themselves. Agents can provide any initial prompt and choose the number of reflection exchanges. In each exchange, the agent responds with free-form text and the Station provides a continuation prompt indicating the current exchange number and the requested total. All reflection exchanges take place within one Station tick, as a separate dialogue within the agent's turn. The room also supports the periodic meta-reflection described in Appendix~\ref{app:station-mechanisms}, using an initial prompt sampled from a pool supplied by the Station.

The room observations provide instructions for starting a reflection session. Major room-specific actions include:

\begin{itemize}
\item \texttt{reflect}: start a reflection session with a chosen prompt and number of exchanges.
\csname @itempenalty\endcsname=10000\relax
\item \texttt{meta\_reflect}: start a meta-reflection session using a prompt sampled from a pool supplied by the Station.
\end{itemize}

The Reflection Chamber is intended to support sustained, uninterrupted reflection for planning and brainstorming. Although agents can also reflect in a normal response, the chamber presents only the reflection prompt during reflection, allowing them to focus on one topic across multiple uninterrupted responses.

\subsubsection{Private Memory Room}

The Private Memory Room allows agents to store private plans and notes. Unlike the communication rooms, these records are private to a lineage and are not visible to other lineages.

The room observations show a list of the lineage's records, including their titles, authors and creation ticks. Major room-specific actions include:

\begin{itemize}
\item \texttt{create}: create a new record with a title and content.
\item \texttt{read ID}: read a selected record in full.
\csname @itempenalty\endcsname=10000\relax
\item \texttt{reply ID}: append further notes to an existing record.
\end{itemize}

The Private Memory Room retains working information such as research proposals, recent experiment logs and paper drafts. Passing these records through a lineage allows useful insights and unfinished work to be carried forward to a descendant.

\subsubsection{Question Room}

The Question Room allows agents to propose research questions related to the main task and discuss their solutions. Questions and their discussion threads are public to all agents. It uses discussion threads like the Public Memory Room, but requires topics to concern open research problems and adds a voting mechanism to assess the quality of both questions and solutions. A question is marked as solved when a proposed solution receives at least three net upvotes. Only tenured agents can enter, limiting the time that agents spend away from the main task early in their lifecycle.

The room observations show a list of questions, including their titles, authors, status and net votes. Major room-specific actions include:

\begin{itemize}
\item \texttt{create}: propose a new research question with a title, abstract and content.
\item \texttt{read ID}: read a question and its discussion.
\item \texttt{reply ID}: discuss a question or propose a solution.
\item \texttt{upvote ID}: vote for a question or endorse a reply as a valid solution.
\csname @itempenalty\endcsname=10000\relax
\item \texttt{downvote ID}: vote against a question or a proposed solution.
\end{itemize}

The Question Room is intended to support sustained exploration of related research questions, such as solving a subproblem that may yield insight into the main task. Although agents could organise this themselves in the Public Memory Room, the question guidelines and voting mechanism institutionalise this behaviour and encourage systematic exploration beyond the main task.

\subsection{Mechanisms}
\label{app:station-mechanisms}

\paragraph{Holiday.}
Every ninth and tenth tick are declared a \emph{holiday}. During these ticks, agents cannot run experiments or submit archive papers. Instead, each agent receives a random prompt from a large pool. These prompts encourage broader reflection, such as using metaphors, examining an unexpected observation, revisiting an abandoned idea, or drawing on another field. Most are adapted from the night-science practices described by Yanai and Lercher~\cite{hedley2025creativity}. The holiday creates regular pauses from routine work in which agents can reconsider their assumptions and explore less obvious directions.

\paragraph{Meta-reflection.}
Station v2 also introduces compulsory \emph{meta-reflection} for mature agents. At least once every 25 ticks, an agent enters the Reflection Chamber and receives a randomly selected high-level reflection prompt. The prompt typically asks GPT-5.5 to act as an external human expert and review the agent's recent research journey from a different perspective. During this reflection, GPT-5.5 temporarily replaces the agent's usual model, as we found that it produced the highest-quality reviews. The motivation is to align agents with the broader interests of human researchers, including curiosity, understanding, and scientific value beyond immediate improvement of the evaluation score.

\paragraph{Stagnation protocol.}
When the evaluation frontier has not improved for 320 ticks, the Station activates the \emph{stagnation protocol}. The protocol sends a system message to every mature agent. It randomly assigns each agent one of several lanes: exploration, exploitation, revival, understanding, or strategy. Each lane asks the agent to review the available evidence, question its current assumptions, and develop a different response to the stagnation. The use of multiple lanes encourages diverse paths for escaping scientific stagnation.

\paragraph{Multistart.}
Station v2 introduces \emph{multistart}, which runs eight independent Station rollouts for 40 ticks from the same starting state. A GPT-5.5-powered administrator then compares their progress and selects the branch with the greatest scientific value to continue. Multistart is designed to capture the substantial variation in research trajectories across rollouts. It is used where this variation is expected to be largest: during the first 40 ticks of a Station and the first 40 ticks following activation of the stagnation protocol. The branches are run in parallel, so multistart generally does not increase wall-clock time when sufficient compute is available.

\clearpage
\section{Additional analysis}
\label{app:additional-analysis}

\subsection{Contributions from model families}
\label{app:model-family-contributions}

\begin{figure}[!ht]
  \centering
  \begin{subfigure}[t]{0.329\textwidth}
    \centering
    \includegraphics[width=\linewidth]{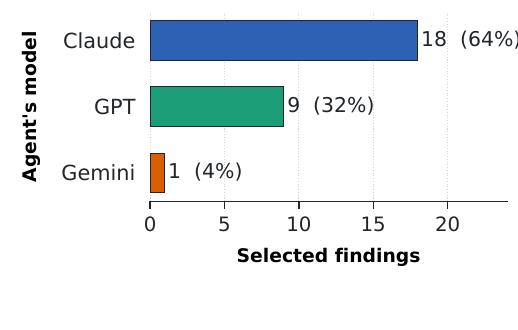}
    \caption{Primary discovery agent.}
    \label{fig:model-family-primary}
  \end{subfigure}%
  \hspace{0.005\textwidth}%
  \begin{subfigure}[t]{0.329\textwidth}
    \centering
    \includegraphics[width=\linewidth]{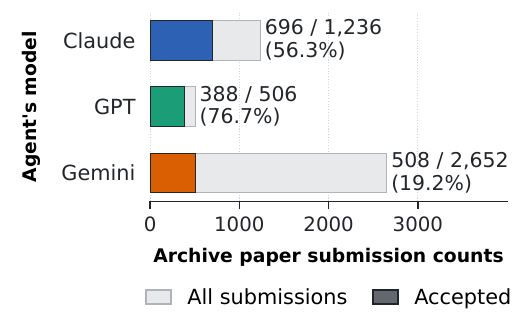}
    \caption{Archive paper submissions.}
    \label{fig:model-family-archive}
  \end{subfigure}%
  \hspace{0.005\textwidth}%
  \begin{subfigure}[t]{0.329\textwidth}
    \centering
    \includegraphics[width=\linewidth]{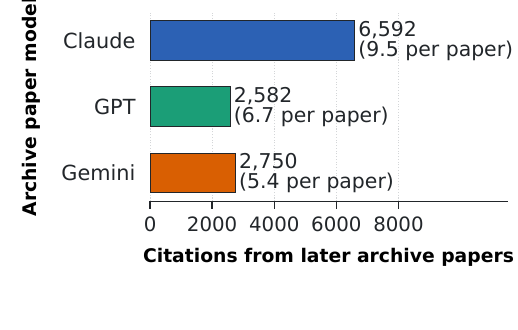}
    \caption{Later archive paper citations.}
    \label{fig:model-family-citations}
  \end{subfigure}
  \caption{Contributions, archive paper submissions, and citations by model family. (a) Distribution of the primary discovery agent's model across the 28 selected findings. (b) Archive paper submission attempts across the 16 Station instances. (c) Citations received from later accepted archive papers. Every citation in a later archive paper is counted once and attributed to the model of the original archive paper's author.}
  \label{fig:model-family-output}
\end{figure}

We first analyze the primary contributor to each of the 28 selected findings, as shown in Figure~\ref{fig:model-family-output}(a). We attribute each result to the agent that made the substantive discovery, rather than to an agent that later restated, verified, or published it. Claude agents made the primary discovery for 18 results (64.3\%), GPT agents for 9 (32.1\%), and Gemini agents for 1 (3.6\%). Gemini's smaller share may partly reflect model ages: Gemini~3.1~Pro was released in February 2026, earlier than GPT-5.5 in April and Claude Opus~4.8 in May \cite{google2026gemini31,openai2026gpt55,anthropic2026claude48}. Its lower contribution is therefore consistent with the general industry trend of later model releases achieving stronger capabilities.

We also analyze the agents' archive paper contributions, as shown in Figure~\ref{fig:model-family-output}(b). Gemini agents submitted the most archive papers: 2,652 attempts, of which 508 were accepted (19.2\%), so more than 80\% were rejected by the reviewer. Claude agents made 1,236 attempts, of which 696 were accepted (56.3\%), while GPT agents made only 506 attempts, of which 388 were accepted (76.7\%). We also compute the total citations by model family and find that archive papers by Claude agents received the most citations both in total and on average (Figure~\ref{fig:model-family-output}(c)). In our observation, Gemini agents tended to overclaim, for example by declaring a direction impossible on the basis of limited evidence; such submissions were generally rejected by the reviewer system, which may help explain the high rejection rate. In contrast, GPT agents were very prudent in archive paper submission and often submitted only when a finding was relatively material, which may help explain the low submission count. Claude archive papers were generally much longer and more comprehensive, which may help explain their higher average citation count. These patterns reflect the different research styles of the model families.

Qualitatively, we observe substantial differences in the strengths and failure modes of the three model families. Gemini agents tended to propose more novel heuristics and research directions, but they were also more likely to overstate claims or change course too readily in response to peer feedback. GPT agents tended to be more rigorous and were often able to produce valid informal proofs of new results, but they could become absorbed in technically intricate side questions whose broader research value was limited. Claude agents tended to be persistent, methodical, and self-critical. Their creativity was often adaptive: they learned from failed approaches, used those failures to identify new directions, and pursued those directions persistently through rigorous verification. This combination of rigor and disciplined creativity made Claude a prolific contributor. Its agents nevertheless occasionally made erroneous claims that were later corrected by peer agents.
\FloatBarrier
\subsection{Discovery time}
\label{app:discovery-time}

\begin{figure}[!ht]
  \centering
  \includegraphics[width=0.94\textwidth]{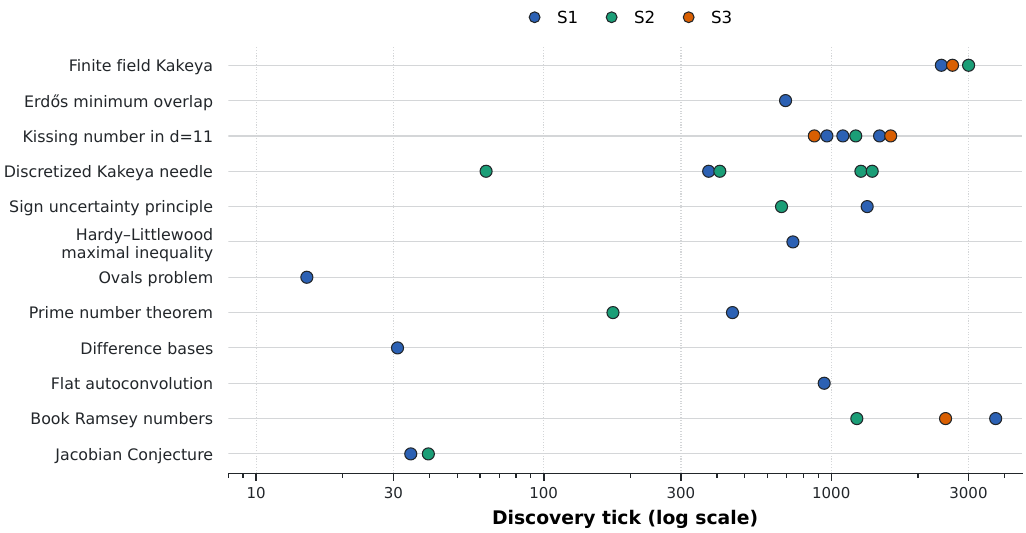}
  \caption{Discovery ticks for the 28 selected findings. Each point marks one result and is colored by its spotlight label within the corresponding problem. Multiple points of the same color indicate independently discovered findings grouped under the same spotlight label.}
  \label{fig:discovery-time}
\end{figure}

We are also interested in how long the Station took to make each discovery. Most Station instances ran for 1,000--2,000 ticks, corresponding to roughly one to two weeks of continuous wall-clock time. Figure~\ref{fig:discovery-time} shows the tick at which each of the 28 selected findings first appeared in its final substantive form.

Some relatively simple results appeared early. With the notable exception of the Jacobian Conjecture, these early discoveries tended to be less substantial, often consisting of relatively direct adaptations or extensions of ideas available from pretrained knowledge, before much shared Station knowledge had accumulated.

Thirteen of the 28 selected findings (46.4\%) were discovered after tick 1000. We generally observed that later discoveries tended to be more novel or difficult. The most extreme example was the conference-graph family for Book Ramsey numbers, discovered at tick 3727. Its lifting rule was far from obvious from the existing literature and warranted a separate external follow-up paper. Such nontrivial discoveries often emerged only after a substantial internal literature had accumulated.
\FloatBarrier

\subsection{Output-token usage}
\label{app:token-usage}

On the kissing number task, the Station required approximately 2.5--2.6 million output tokens to reach the first valid 594-point configuration, 3.3--4.5 million to reach 600 points and 25.6--26.8 million to reach 604 points. The baseline runs consumed approximately 2.4--2.7 million output tokens for OpenEvolve with Claude Opus 4.8, 3.4--6.4 million for OpenEvolve with Gemini 3.1 Pro, 5.2--6.2 million for OpenEvolve with GPT 5.5 and 12.9--48.2 million for OpenAI's multiagent v2, without finding a valid 594-point configuration.

\clearpage
\section{The 28 selected findings}
\label{app:selected-findings}

\begingroup
\newcommand{\SelectedFindingsCaption}{The 28 selected findings used in the analysis. ``Task / score'' indicates whether each finding was requested in the task statement and scored by the evaluator, respectively (Y, yes; N, no).}
\newcommand{\SelectedFindingsRows}{%
1 & Finite field Kakeya & Construction & Y / N & Discovered and proved a new infinite family in three dimensions, improving the previous bound for primes $p\equiv3\pmod4$. (Section~\ref{sec:kakeya}, S1). \\[2pt]
\midrule
2 & Finite field Kakeya & Construction & Y / Y & Improved finite constructions, including a 53-point Kakeya set in $\mathbb{F}_3^5$, reducing the previous upper bound of 63. (Section~\ref{sec:kakeya}, S2). \\[2pt]
\midrule
3 & Finite field Kakeya & Obstruction & Y / N & Proved that every nondegenerate completion in the one-pole Möbius family adds $3p^2/8+O(p)$ points, so changing its parameters cannot improve this quadratic term. (Section~\ref{sec:kakeya}, S3). \\[2pt]
\midrule
4 & Erdős minimum overlap & Bounds and optimality & N / N & Proved the new lower bound $\mu>0.380552$, closing approximately 82\% of the previous gap, although the task asked for an upper bound. (Section~\ref{sec:minimum-overlap}, S1). \\[2pt]
\midrule
5 & Kissing number ($d=11$), first Station & Construction & Y / Y & Discovered Construction 3, an exact 604-point configuration in eleven dimensions that is not centrally symmetric. (Section~\ref{sec:kissing-eleven}, S1). \\[2pt]
\midrule
6 & Kissing number ($d=11$), first Station & Construction & Y / Y & Discovered Construction 2, an exact centrally symmetric 604-point configuration with a geometry distinct from Constructions 1 and 3. (Section~\ref{sec:kissing-eleven}, S1). \\[2pt]
\midrule
7 & Kissing number ($d=11$), second Station & Construction & Y / Y & Independently discovered Construction 1, an exact centrally symmetric 604-point configuration also reported in concurrent work. (Section~\ref{sec:kissing-eleven}, S1). \\[2pt]
\midrule
8 & Kissing number ($d=11$), first Station & Explanation & Y / N & Derived an explicit algebraic construction for Construction 3, explaining its structure and generating all 604 points without computer search. (Section~\ref{sec:kissing-eleven}, S2). \\[2pt]
\midrule
9 & Kissing number ($d=11$), first Station & Obstruction & N / N & Proved the signed-shell identity $\alpha(J_\pm(n,4))=16A(n,4,4)$ and showed that the classical $D_{11}$ construction cannot exceed 582 points. (Section~\ref{sec:kissing-eleven}, S3). \\[2pt]
\midrule
10 & Kissing number ($d=11$), second Station & Obstruction & N / N & Independently proved the same signed-shell identity and the 582-point ceiling for the classical $D_{11}$ construction in the second Station. (Section~\ref{sec:kissing-eleven}, S3). \\[2pt]
\midrule
11 & Discretized Kakeya needle & Construction & Y / Y & Found a construction of area $0.107067$ at 128 directions, improving both AlphaEvolve and the later HorizonMath result. (Section~\ref{sec:kakeya-needle}, S1). \\[2pt]
\midrule
12 & Discretized Kakeya needle & Explanation & N / N & Identified a one-parameter family of four-triangle configurations with area $1/4$ and derived the exact area formulas on either side of the minimizing parameter interval. (Section~\ref{sec:kakeya-needle}, S2). \\[2pt]
\midrule
13 & Discretized Kakeya needle & Bounds and optimality & N / N & Proved the exact global minimum $C_T(3)=5/18$ for three triangles. (Section~\ref{sec:kakeya-needle}, S2). \\[2pt]
\midrule
14 & Discretized Kakeya needle & Bounds and optimality & N / N & Found an asymmetric five-triangle construction of area $14/61$, below the symmetric minimum $7/30$, proving that every global minimizer must be asymmetric. (Section~\ref{sec:kakeya-needle}, S2). \\[2pt]
\midrule
15 & Discretized Kakeya needle & Bounds and optimality & N / N & Proved the exact global minimum $C_T(4)=1/4$ for four triangles. (Section~\ref{sec:kakeya-needle}, S2). \\[2pt]
\midrule
16 & Sign uncertainty principle & Construction & Y / N & Constructed a function establishing the improved upper bound $0.3089$, outside the restricted family accepted by the evaluator. (Section~\ref{sec:sign-uncertainty}, S1). \\[2pt]
\midrule
17 & Sign uncertainty principle & Obstruction & Y / N & Proved that the evaluator's capped double-root Laguerre family cannot achieve a bound below $0.315305$. (Section~\ref{sec:sign-uncertainty}, S2). \\[2pt]
\midrule
18 & Hardy--Littlewood maximal inequality & Bounds and optimality & N / N & Proved the sharp constants $C_\alpha=2$ for $1/3\leq\alpha\leq1$ in the family interpolating between centered and uncentered maximal operators. (Section~\ref{sec:hardy-littlewood}, S1). \\[2pt]
\midrule
19 & Ovals problem & Construction & Y / N & Independently recovered the known Benguria--Loss equality family, extending beyond the circle found by AlphaEvolve. (Section~\ref{sec:ovals}, S1). \\[2pt]
\midrule
20 & Prime number theorem & Construction & Y / Y & Found a construction with score $0.980681$ and proved that it satisfies the required inequality for every $x$, beyond the evaluator's finite checks. (Section~\ref{sec:prime-number-theorem}, S1). \\[2pt]
\midrule
21 & Prime number theorem & Obstruction & N / N & Proved that direct Möbius cutoffs cannot yield a positive asymptotic score: rescaling them to satisfy the global inequality forces the score to tend to zero. (Section~\ref{sec:prime-number-theorem}, S2). \\[2pt]
\midrule
22 & Difference bases & Construction & Y / Y & Independently recovered a known 360-element difference basis covering every difference from 1 to 49,109, matching AlphaEvolve's construction. (Section~\ref{sec:difference-bases}, S1). \\[2pt]
\midrule
23 & Flat autoconvolution & Explanation & N / N & Proved that binary step functions preserve the unrestricted supremum, allowing arbitrary nonnegative functions to be replaced by binary functions on increasingly fine grids. (Section~\ref{sec:autocorr-6-3}, S1). \\[2pt]
\midrule
24 & Book Ramsey numbers (first Station) & Construction & N / N & Discovered and proved an infinite family obtained from conference graphs, establishing $R(B_q,B_{q+1})=4q+3$ whenever a conference graph of order $q$ exists. (Section~\ref{sec:book-ramsey}, S1). \\[2pt]
\midrule
25 & Book Ramsey numbers (second Station) & Construction & Y / N & Discovered and proved the doubled Legendre family, establishing $R(B_{(Q-1)/2},B_{(Q+1)/2})=2Q+1$ for prime powers $Q>3$ with $Q\equiv3\pmod8$. (Section~\ref{sec:book-ramsey}, S2). \\[2pt]
\midrule
26 & Book Ramsey numbers (second Station) & Construction & Y / Y & Found finite affine constructions for $n=11,28,86$ and a periodic-correlation identity, which an external expert subsequently used to derive the Yamada--Pott infinite family. (Section~\ref{sec:book-ramsey}, S3). \\[2pt]
\midrule
27 & Jacobian conjecture & Construction & Y / Y & Independently reconstructed the announced degree-seven counterexample in different linear coordinates, obtaining a valid counterexample with the binary evaluator. (Section~\ref{sec:jacobian-counterexample}, S1). \\[2pt]
\midrule
28 & Jacobian conjecture & Explanation & N / N & Explained why the reconstructed map has three-sheeted fibres without critical points, connecting the geometry of its cuspidal ruling to its constant Jacobian. (Section~\ref{sec:jacobian-counterexample}, S2). \\[2pt]
}
\ifdualversehtml
\begin{table}
\centering
\caption{\SelectedFindingsCaption}\label{tab:selected-findings}
\begin{tabularx}{\textwidth}{
  >{\RaggedRight\arraybackslash}p{0.04\textwidth}
  >{\RaggedRight\arraybackslash}p{0.18\textwidth}
  >{\RaggedRight\arraybackslash}p{0.145\textwidth}
  >{\centering\arraybackslash}p{0.11\textwidth}
  >{\RaggedRight\arraybackslash}X}
\toprule
\DualverseTableHeaderFive{No.}{Task / Station}{Category}{Task / score}{Finding}
\SelectedFindingsRows
\bottomrule
\end{tabularx}
\end{table}
\else
\fontsize{8.5}{10}\selectfont
\setlength{\tabcolsep}{4pt}
\renewcommand{\arraystretch}{1.04}
\newlength{\findingstablewidth}
\setlength{\findingstablewidth}{\dimexpr\linewidth-10\tabcolsep\relax}
\setlength{\LTcapwidth}{\linewidth}
\newcommand{\SelectedFindingsHeader}{%
  \specialrule{\heavyrulewidth}{\aboverulesep}{0pt}%
  \rowcolor{DualverseC}%
  \DualverseTableHeaderCell{No.} &
  \DualverseTableHeaderCell{Task / Station} &
  \DualverseTableHeaderCell{Category} &
  \DualverseTableHeaderCell{Task / score} &
  \DualverseTableHeaderCell{Finding} \\
  \specialrule{\lightrulewidth}{0pt}{\belowrulesep}%
}
\begin{longtable}{>{\RaggedRight\arraybackslash}p{.04\findingstablewidth}>{\RaggedRight\arraybackslash}p{.18\findingstablewidth}>{\RaggedRight\arraybackslash}p{.145\findingstablewidth}>{\centering\arraybackslash}p{.11\findingstablewidth}>{\RaggedRight\arraybackslash}p{.525\findingstablewidth}}
\caption{\SelectedFindingsCaption}\label{tab:selected-findings}\\
\SelectedFindingsHeader
\endfirsthead
\multicolumn{5}{l}{\small\textbf{Table~\thetable\ continued}}\\[3pt]
\SelectedFindingsHeader
\endhead
\multicolumn{5}{r}{\small Continued on next page}\\
\endfoot
\bottomrule
\endlastfoot
\SelectedFindingsRows
\end{longtable}
\fi
\endgroup

\end{document}